\documentclass{article}

\usepackage{amsmath,amsfonts,bm}

\def\eqref#1{equation~\ref{#1}}

\def\1{\bm{1}}

\def\vc{{\bm{c}}}
\def\vd{{\bm{d}}}

\def\vx{{\bm{x}}}
\def\vy{{\bm{y}}}
\def\vz{{\bm{z}}}

\DeclareMathAlphabet{\mathsfit}{\encodingdefault}{\sfdefault}{m}{sl}
\SetMathAlphabet{\mathsfit}{bold}{\encodingdefault}{\sfdefault}{bx}{n}
\newcommand{\tens}[1]{\bm{\mathsfit{#1}}}

\def\tG{{\tens{G}}}

\def\sI{{\mathbb{I}}}

\def\sL{{\mathbb{L}}}

\def\sR{{\mathbb{R}}}

\usepackage{microtype}
\usepackage{graphicx}
\usepackage{subfigure}
\usepackage{booktabs}

\usepackage{multirow}
\usepackage{multicol}

\usepackage{url}
\usepackage{doi}

\usepackage{stfloats} 

\usepackage{xcolor}

\newcommand{\best}[1]{\textbf{\textcolor{green!25!black}{\underline{#1}}}}
\newcommand{\bestErr}[1]{\textbf{\textcolor{green!25!black}{#1}}}
\newcommand{\bad}[1]{\textit{\textcolor{red!40!black}{#1}}}

\usepackage[accepted]{icml2020}

\icmltitlerunning{Learning Similarity Metrics for Numerical Simulations}

\begin{document}

\twocolumn[
\icmltitle{Learning Similarity Metrics for Numerical Simulations}

\icmlsetsymbol{equal}{*}

\begin{icmlauthorlist}
\icmlauthor{Georg Kohl}{tum}
\icmlauthor{Kiwon Um}{tum}
\icmlauthor{Nils Thuerey}{tum}
\end{icmlauthorlist}

\icmlaffiliation{tum}{Department of Informatics, Technical University of Munich, Munich, Germany}

\icmlcorrespondingauthor{Georg Kohl}{georg.kohl@tum.de}

\icmlkeywords{metric learning, CNNs, PDEs, numerical simulation, perceptual evaluation, physics simulation}

\vskip 0.3in
]

\begin{NoHyper}
\printAffiliationsAndNotice{}  
\end{NoHyper}

\begin{abstract}
 We propose a neural network-based approach that computes a stable and generalizing metric (\textit{LSiM}) to compare data from a variety of numerical simulation sources. We focus on scalar time-dependent 2D data that commonly arises from motion and transport-based partial differential equations (PDEs). Our method employs a Siamese network architecture that is motivated by the mathematical properties of a metric. We leverage a controllable data generation setup with PDE solvers to create increasingly different outputs from a reference simulation in a controlled environment. A central component of our learned metric is a specialized loss function that introduces knowledge about the correlation between single data samples into the training process. To demonstrate that the proposed approach outperforms existing metrics for vector spaces and other learned, image-based metrics, we evaluate the different methods on a large range of test data. Additionally, we analyze generalization benefits of an adjustable training data difficulty and demonstrate the robustness of \textit{LSiM} via an evaluation on three real-world data sets.
\end{abstract}

\section{Introduction} \label{sec: intro}
Evaluating computational tasks for complex data sets is a fundamental problem in all computational disciplines. Regular vector space metrics, such as the $L^2$ distance, were shown to be very unreliable \citep{wang2004,zhang2018},
and the advent of deep learning techniques with convolutional neural networks (CNNs) made it possible to more reliably evaluate complex data domains such as natural images, texts \citep{benajiba2018}, or speech \citep{wang2018}.
Our central aim is to demonstrate the usefulness of CNN-based evaluations in the context of numerical simulations. These simulations are the basis for a wide range of applications ranging from blood flow simulations to aircraft design. Specifically, we propose a novel learned simulation metric (\textit{LSiM}) that allows for a reliable similarity evaluation of simulation data.

Potential applications of such a metric arise in all areas where numerical simulations are performed or similar data is gathered from observations. For example, accurate evaluations of existing and new simulation methods with respect to a known ground truth solution \citep{oberkampf2004} can be performed more reliably than with a regular vector norm. Another good example is weather data for which complex transport processes and chemical reactions make in-place comparisons with common metrics unreliable \citep{jolliffe2012}. Likewise, the long-standing, open questions of turbulence \citep{moin1998,lin1998} can benefit from improved methods for measuring the similarity and differences in data sets and observations.

\begin{figure*}[ht]
    \centering
    \includegraphics[width=1.0\textwidth]{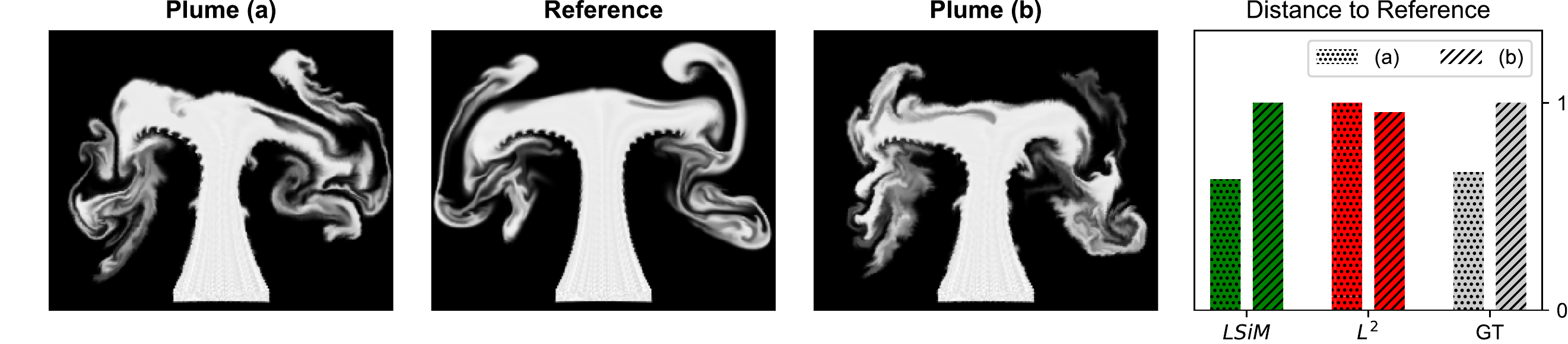}
    \vspace{-0.8cm}
    \caption{Example of field data from a fluid simulation of hot smoke with normalized distances for different metrics. Our method (\textit{LSiM}, green) approximates the ground truth distances (GT, gray) determined by the data generation method best, i.e., version (a) is closer to the ground truth data than (b). An $L^2$ metric (red) erroneously yields a reversed ordering.}
    \label{fig: visual similarity}
\end{figure*}

In this work, we focus on field data, i.e., dense grids of scalar values, similar to images, which were generated with known partial differential equations (PDEs) in order to ensure the availability of ground truth solutions. While we focus on 2D data in the following to make comparisons with existing techniques from imaging applications possible, our approach naturally extends to higher dimensions.
Every sample of this 2D data can be regarded a high dimensional vector, so metrics on the corresponding vector space are applicable to evaluate similarities. These metrics, in the following denoted as \emph{shallow metrics}, are typically simple, element-wise functions such as $L^1$ or $L^2$ distances. Their inherent problem is that they cannot compare structures on different scales or contextual information.

Many practical problems require solutions over time and need a vast number of non-linear operations that often result in substantial changes of the solutions even for small changes of the inputs. Hence, despite being based on known, continuous formulations, these systems can be seen as \emph{chaotic}.
We illustrate this behavior in Fig.~\ref{fig: visual similarity}, where two smoke flows are compared to a reference simulation. A single simulation parameter was varied for these examples, and a visual inspection shows that smoke plume (a) is more similar to the reference. This matches the data generation process: version (a) has a significantly smaller parameter change than (b) as shown in the inset graph on the right.
\textit{LSiM} robustly predicts the ground truth distances while the $L^2$ metric labels plume (b) as more similar. In our work, we focus on retrieving the relative distances of simulated data sets. Thus, we do not aim for retrieving the absolute parameter change but a relative distance that preserves ordering with respect to this parameter.

Using existing image metrics based on CNNs for this problem is not optimal either: natural images only cover a small fraction of the space of possible 2D data, and numerical simulation outputs are located in a fundamentally different data manifold within this space. Hence, there are crucial aspects that cannot be captured by purely learning from photographs. Furthermore, we have full control over the data generation process for simulation data. 
As a result, we can create arbitrary amounts of training data with gradual changes and a ground truth ordering.
With this data, we can learn a metric that is not only able to directly extract and use features but also encodes interactions between them.
The central contributions of our work are as follows:
\vspace{-0.2cm}
\begin{itemize}
    \setlength\itemsep{0.0cm}
    \item We propose a Siamese network architecture with feature map normalization, which is able to learn a metric that generalizes well to unseen motion and transport-based simulation methods.
    \item We propose a novel loss function that combines a correlation loss term with a mean squared error to improve the accuracy of the learned metric.
    \item In addition, we show how a data generation approach for numerical simulations can be employed to train networks with general and robust feature extractors for metric calculations.
\end{itemize}
\vspace{-0.2cm}
Our source code, data sets, and final model are available at \url{https://github.com/tum-pbs/LSIM}.


\section{Related Work} \label{sec: related work}
One of the earliest methods to go beyond using simple metrics based on $L^p$ norms for natural images was the structural similarity index \citep{wang2004}. Despite improvements, this method can still be considered a shallow metric.
Over the years, multiple large databases for human evaluations of natural images were presented, for instance, CSIQ \citep{larson2010}, TID2013 \citep{ponomarenko2015}, and CID:IQ \citep{liu2014}. With this data and the discovery that CNNs can create very powerful feature extractors that are able to recognize patterns and structures, deep feature maps quickly became established as means for evaluation
\citep{amirshahi2016, berardino2017, bosse2016, kang2014, kim2017}. Recently, these methods were improved by predicting the distribution of human evaluations instead of directly learning distance values \citep{prashnani2018, talebi2018b}. \citeauthor{zhang2018} compared different architecture and levels of supervision, and showed that metrics can be interpreted as a transfer learning approach by applying a linear weighting to the feature maps of any network architecture to form the image metric \textit{LPIPS v0.1}. Typical use cases of these image-based CNN metrics are computer vision tasks such as detail enhancement \citep{talebi2018a}, style transfer, and super-resolution \citep{johnson2016}.
Generative adversarial networks also leverage CNN-based losses by training a discriminator network in parallel to the generation task \citep{dosovitskiy2016}.
\vspace{-0.02cm}

Siamese network architectures are known to work well for a variety of comparison tasks such as audio \citep{zhang2017}, satellite images \citep{he2019}, or the similarity of interior product designs \citep{bell2015}. Furthermore, they yield robust object trackers \citep{bertinetto2016}, algorithms for image patch matching \citep{hanif2019}, and for descriptors for fluid flow synthesis \citep{chu2017}.
Inspired by these studies, we use a similar Siamese neural network architecture for our metric learning task.
In contrast to other work on self-supervised learning that utilizes spatial or temporal changes to learn meaningful representations \citep{agrawal2015, wang2015}, our method does not rely on tracked keypoints in the data.
\vspace{-0.02cm}

While correlation terms have been used for learning joint representations by maximizing correlation of projected views \citep{chandar2016} and are popular for style transfer applications via the Gram matrix \citep{ruder2016},
they were not used for learning distance metrics. As we demonstrate below, they can yield significant improvements in terms of the inferred distances.

Similarity metrics for numerical simulations are a topic of ongoing investigation. A variety of specialized metrics have been proposed to overcome the limitations of $L^p$ norms, such as the displacement and amplitude score from the area of weather forecasting \citep{keil2009} as well as permutation based metrics for energy consumption forecasting \citep{haben2014}. Turbulent flows, on the other hand, are often evaluated in terms of aggregated frequency spectra \citep{pitsch2006}. Crowd-sourced evaluations based on the human visual system were also proposed to evaluate simulation methods for physics-based animation \citep{um2017} and for comparing non-oscillatory discretization schemes \citep{um2019}. 
These results indicate that visual evaluations in the context of field data are possible and robust, but they require extensive (and potentially expensive) user studies. Additionally, our method naturally extends to higher dimensions, while human evaluations inherently rely on projections with at most two spatial and one time dimension.


\section{Constructing a CNN-based Metric} \label{sec: lsim}

\begin{figure*}[bp]
    \centering
    \includegraphics[width=1.0\textwidth]{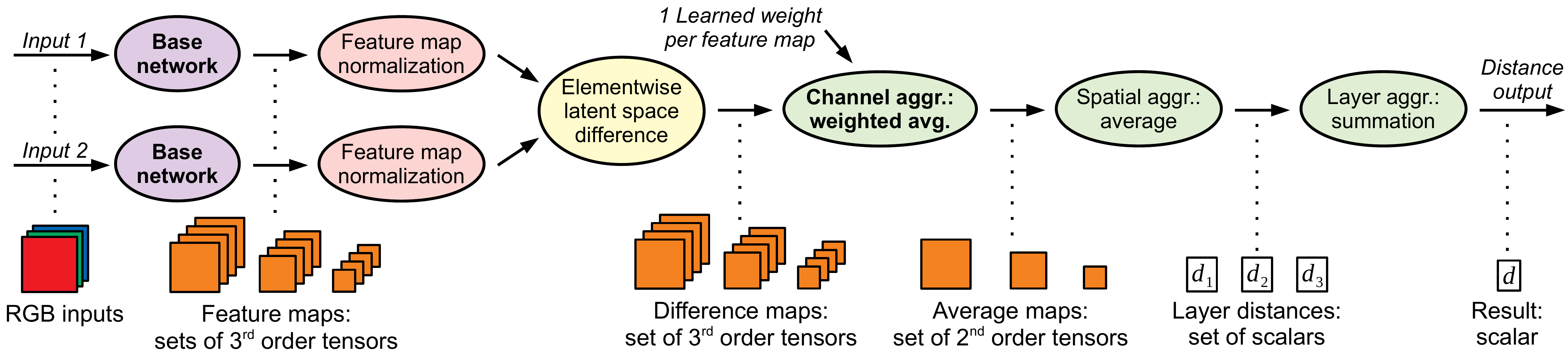}
    \vspace{-0.7cm}
    \caption{Overview of the proposed distance computation for a simplified base network that contains three layers with four feature maps each in this example. The output shape for every operation is illustrated below the transitions in orange and white. Bold operations are learned, i.e., contain weights influenced by the training process.}
    \label{fig: distance computation}
\end{figure*}

In the following, we explain our considerations when employing CNNs as evaluation metrics. For a comparison that corresponds to our intuitive understanding of distances, an underlying \emph{metric} has to obey certain criteria. More precisely, a function $m : \sI \times \sI \to [0, \infty)$ is a metric on its input space $\sI$ if it satisfies the following properties $\forall \vx,\vy,\vz \in \sI$:
\begin{align}
    m(\vx,\vy) \; &\geq \; 0                && \text{non-negativity} \label{eq: NonNeg}\\
    m(\vx,\vy) \; &= \; m(\vy,\vx)              && \text{symmetry} \label{eq: Sym}\\
    m(\vx,\vy) \; &\leq \; m(\vx,\vz) + m(\vz,\vy)  && \text{triangle ineq.} \label{eq: TriIneq}\\
    m(\vx,\vy) \; &= 0 \; \iff \; \vx = \vy     && \text{identity of indisc.} \label{eq: IoI}
\end{align}
The properties (\ref{eq: NonNeg}) and (\ref{eq: Sym}) are crucial as distances should be symmetric and have a clear lower bound. Eq.~(\ref{eq: TriIneq}) ensures that direct distances cannot be longer than a detour. Property (\ref{eq: IoI}), on the other hand, is not really useful for discrete operations as approximation errors and floating point operations can easily lead to a distance of zero for slightly different inputs. Hence, we focus on a relaxed, more meaningful definition $m(\vx,\vx) = 0 \; \forall \vx \in \sI$, which leads to a so-called \emph{pseudometric}. It allows for a distance of zero for different inputs but has to be able to spot identical inputs. 

We realize these requirements for a pseudometric with an architecture that follows popular perceptual metrics such as \textit{LPIPS}: The activations of a CNN are compared in latent space, accumulated with a set of weights, and the resulting per-feature distances are aggregated to produce a final distance value. Fig.~\ref{fig: distance computation} gives a visual overview of this process.

To show that the proposed Siamese architecture by construction qualifies as a pseudometric, the function 
\begin{equation*}
    m(\vx,\vy) \;=\; m_2( m_1(\vx), m_1(\vy) )
\end{equation*}
computed by our network is split into two parts: $m_1 : \sI \to \sL$ to compute the latent space embeddings $\tilde{\vx} = m_1(\vx), \tilde{\vy} = m_1(\vy)$ from each input, and $m_2 : \sL \to [0, \infty)$ to compare these points in the latent space $\sL$. We chose operations for $m_2$ such that it forms a metric $\forall \tilde{\vx}, \tilde{\vy} \in \sL$. Since $m_1$ always maps to $\sL$, this means $m$ has the properties (\ref{eq: NonNeg}), (\ref{eq: Sym}), and (\ref{eq: TriIneq}) on $\sI$ for any possible mapping $m_1$, i.e., only a metric on $\sL$ is required. To achieve property (\ref{eq: IoI}), $m_1$ would need to be injective, but the compression of typical feature extractors precludes this. However, if $m_1$ is deterministic $m(\vx, \vx) = 0 \; \forall \vx \in \sI$ is still fulfilled since identical inputs result in the same point in latent space and thus a distance of zero. More details for this proof can be found in App.~\ref{append: metric properties}.

\subsection{Base Network}
The sole purpose of the base network (Fig.~\ref{fig: distance computation}, in purple) is to extract feature maps from both inputs. The Siamese architecture implies that the weights of the base network are shared for both inputs, meaning all feature maps are comparable. We experimented with the feature extracting layers from various CNN architectures, such as AlexNet \citep{krizhevsky2017}, VGG \citep{simonyan2015}, SqueezeNet \citep{iandola2016}, and a fluid flow prediction network \citep{thuerey2018}. We considered three variants of these networks: using the original pre-trained weights, fine-tuning them, or re-training the full networks from scratch. In contrast to typical CNN tasks where only the result of the final output layer is further processed, we make use of the full range of extracted features across the layers of a CNN (see Fig.~\ref{fig: distance computation}). This implies a slightly different goal compared to regular training: while early features should be general enough to allow for extracting more complex features in deeper layers, this is not their sole purpose. Rather, features in earlier layers of the network can directly participate in the final distance calculation and can yield important cues.

We achieved the best performance for our data sets using a base network architecture with five layers, similar to a reduced AlexNet, that was trained from scratch (see App.~\ref{append: base network}). This feature extractor is fully convolutional and thus allows for varying spatial input dimensions, but for comparability to other models we keep the input size constant at $224\times224$ for our evaluation. In separate tests with interpolated inputs, we found that the metric still works well for scaling factors in the range $[0.5, 2]$. 

\subsection{Feature Map Normalization} \label{subsec: feature map norm}
The goal of normalizing the feature maps (Fig.~\ref{fig: distance computation}, in red) is to transform the extracted features of each layer, which typically have very different orders of magnitude, into comparable ranges. While this task could potentially be performed by the learned weights, we found the normalization to yield improved performance in general.

Let $\tG$ denote a 4\textsuperscript{th} order feature tensor with dimensions $(g_b, g_c, g_x, g_y)$ from one layer of the base network. We form a series $\tG_0, \tG_1, \dotsc$ for every possible content of this tensor across our training samples. The normalization only happens in the channel dimension, so all following operations accumulate values along the dimension of $g_c$ while keeping $g_b$, $g_x$, and $g_y$ constant, i.e., are applied independently of the batch and spatial dimensions.
The unit length normalization proposed by \citeauthor{zhang2018}, i.e.,
\begin{equation*}
    \text{norm}_{\text{unit}}(\tG) = \tG \, / \left\| \tG \right\|_2,
\end{equation*}
only considers the current sample. In this case, $\left\| \tG \right\|_2$ is a 3\textsuperscript{rd} order tensor with the Euclidean norms of $\tG$ along the channel dimension. Effectively, this results in a cosine distance, which only measures angles of the latent space vectors. To consider the vector magnitude, the most basic idea is to use the maximum norm of other training samples, and this leads to a global unit length normalization
\begin{equation*}
    \text{norm}_{\text{global}}(\tG) = \tG \, / \, \text{max} \left(  \left\| \tG_0 \right\|_2,  \left\| \tG_1 \right\|_2, \dotsc  \right).
\end{equation*}
Now, the magnitude of the current sample can be compared to other feature vectors, but this is not robust since the largest feature vector could be an outlier with respect to the typical content.
Instead, we individually transform each component of a feature vector with dimension $g_c$ to a standard normal distribution. This is realized by subtracting the mean and dividing by the standard deviation of all features element-wise along the channel dimension as follows:
\begin{equation*}
    \text{norm}_{\text{dist}}(\tG) \;=\; \frac{1}{\sqrt{g_c - 1}} \, \frac{\tG - \text{mean} \left(  \tG_0, \tG_1, \dotsc  \right)}
    {\text{std} \left(  \tG_0, \tG_1, \dotsc  \right)}.
\end{equation*}
These statistics are computed via a preprocessing step over the training data and stay fixed during training, as we did not observe significant improvements with more complicated schedules such as keeping a running mean.
The magnitude of the resulting normalized vectors follows a chi distribution with $k=g_c$ degrees of freedom, but computing its mean $\sqrt{2} \; \Gamma((k+1)/2) \;/\; \Gamma(k/2)$ is expensive\footnote{$\Gamma$ denotes the gamma function for factorials}, especially for larger $k$. Instead, the mode of the chi distribution $\sqrt{g_c - 1}$ that closely approximates its mean is employed to achieve a consistent average magnitude of about one independently of $g_c$. As a result, we can measure angles for the latent space vectors and compare their magnitude in the global length distribution across all layers.

\subsection{Latent Space Differences} 
Computing the difference of two latent space representations $\tilde{\vx}, \tilde{\vy} \in \sL$ that consist of all extracted features from the two inputs $\vx,\vy \in \sI$ lies at the core of the metric. This difference operator in combination with the following aggregations has to ensure that the metric properties above are upheld with respect to $\sL$. Thus, the most obvious approach to employ an element-wise difference $\tilde{\vx_i}-\tilde{\vy_i} \;\forall i \in \left\{ 0,1,\dotsc,\text{dim}(\sL) \right\}$ is not suitable, as it invalidates non-negativity and symmetry.
Instead, exponentiation of an absolute difference via $\lvert \tilde{\vx_i}-\tilde{\vy_i} \rvert^p$ yields an $L^p$ metric on $\sL$, when combined with the correct aggregation and a $p$th root. $\lvert \tilde {\vx_i}-\tilde{\vy_i} \rvert^2$ is used to compute the difference maps (Fig.~\ref{fig: distance computation}, in yellow), as we did not observe significant differences for other values of $p$.

Considering the importance of comparing the extracted features, this simple feature difference does not seem optimal. Rather, one can imagine that improvements in terms of comparing one set of feature activations could lead to overall improvements for derived metrics. We investigated replacing these operations with a pre-trained CNN-based metric for each feature map. This creates a recursive process or ``meta-metric'' that reformulates the initial problem of learning input similarities in terms of learning feature space similarities.
However, as detailed in App.~\ref{append: meta metric}, we did not find any substantial improvements with this recursive approach. This implies that once a large enough number of expressive features is available for comparison, the in-place difference of each feature is sufficient to compare two inputs.

\subsection{Aggregations}
The subsequent aggregation operations (Fig.~\ref{fig: distance computation}, in green) are applied to the difference maps to compress the contained per feature differences along the different dimensions into a single distance value. 
A simple summation in combination with an absolute difference $\lvert \tilde{\vx_i}-\tilde{\vy_i} \rvert$ above leads to an $L^1$ distance on the latent space $\sL$. Similarly, we can show that average or learned weighted average operations are applicable too (see App.~\ref{append: metric properties}). In addition, using a $p$-th power for the latent space difference requires a corresponding root operation after all aggregations, to ensure the metric properties with respect to $\sL$.

To aggregate the difference maps along the channel dimension, we found the weighted average proposed by \citeauthor{zhang2018} to work very well. Thus, we use one learnable weight to control the importance of a feature. The weight is a multiplier for the corresponding difference map before summation along the channel dimension, and is clamped to be non-negative. A negative weight would mean that a larger difference in this feature produces a smaller overall distance, which is not helpful. For regularization, the learned aggregation weights utilize dropout during training, i.e., are randomly set to zero with a probability of 50\%. This ensures that the network cannot rely on single features only, but has to consider multiple features for a more stable evaluation.

For spatial and layer aggregation, functions such as a summation or averaging are sufficient and generally interchangeable.
We experimented with more intricate aggregation functions, e.g., by learning a spatial average or determining layer importance weights dynamically from the inputs. When the base network is fixed and the metric only has very few trainable weights, this did improve the overall performance. But, with a fully trained base network, the feature extraction seems to automatically adopt these aspects making a more complicated aggregation unnecessary.


\section{Data Generation and Training} \label{sec: data}
Similarity data sets for natural images typically rely on changing already existing images with distortions, noise, or other operations and assigning ground truth distances according to the strength of the operation. Since we can control the data creation process for numerical simulations directly, we can generate large amounts of simulation data with increasing dissimilarities by altering the parameters used for the simulations. 
As a result, the data contains more information about the nature of the problem, i.e., which changes of the data distribution should lead to increased distances, than by applying modifications as a post-process.

\subsection{Data Generation} 
Given a set of model equations, e.g., a PDE from fluid dynamics, typical solution methods consist of a solver that, given a set of boundary conditions, computes discrete approximations of the necessary differential operators. The discretized operators and the boundary conditions typically contain problem dependent parameters, which we collectively denote with $p_0, p_1, \dotsc, p_i, \dotsc$ in the following. We only consider time dependent problems, and our solvers start with initial conditions at $t_0$ to compute a series of time steps $t_1, t_2, \dotsc$ until a target point in time ($t_t$) is reached. At that point, we obtain a reference output field $o_0$ from one of the PDE variables, e.g., a velocity.

\begin{figure}[htb]
    \centering
    \includegraphics[width=0.48\textwidth]{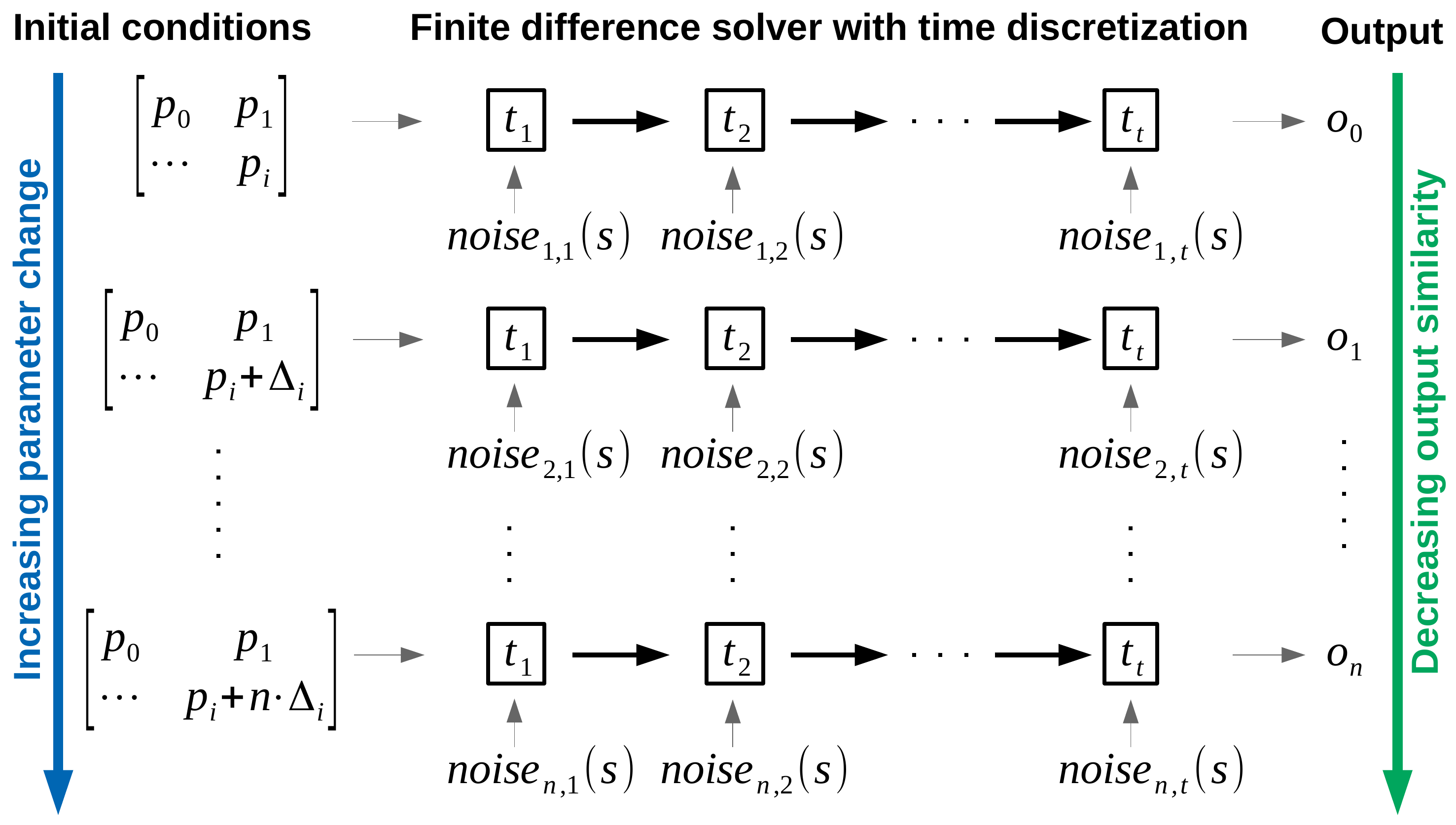}
    \vspace{-0.6cm}
    \caption{General data generation method from a PDE solver for a time dependent problem. With increasing changes of the initial conditions for a parameter $p_i$ in $\Delta_i$ increments, the outputs decrease in similarity. Controlled Gaussian $noise$ is injected in a simulation field of the solver. The difficulty of the learning task can be controlled by scaling $\Delta_i$ as well as the noise variance $v$.
    }
    \label{fig: data generation}
\end{figure}

For data generation, we incrementally change a single parameter $p_i$ in $n$ steps $\Delta_i, 2 \cdot \Delta_i, \dotsc, n \cdot \Delta_i$ to create a series of $n$ outputs $o_1, o_2, \dotsc, o_n$. We consider a series obtained in this way to be increasingly different from $o_0$. To create natural variations of the resulting data distributions, we add Gaussian noise fields with zero mean and adjustable variance $v$ to an appropriate simulation field such as a velocity. This noise allows us to generate a large number of varied data samples for a single simulation parameter $p_i$. Furthermore, $v$ serves as an additional parameter that can be varied in isolation to observe the same simulation with different levels of interference. This is similar in nature to numerical errors introduced by discretization schemes.
These perturbations enlarge the space covered by the training data, and we found that training networks with suitable noise levels improves robustness as we will demonstrate below. The process for data generation is summarized in Fig.~\ref{fig: data generation}.

As PDEs can model extremely complex and chaotic behaviour, there is no guarantee that the outputs always exhibit increasing dissimilarity with the increasing parameter change. This behaviour is what makes the task of similarity assessment so challenging. Even if the solutions are essentially chaotic, their behaviour is not arbitrary but rather governed by the rules of the underlying PDE. 
For our data set, we choose the following range of representative PDEs:
We include a pure Advection-Diffusion model (AD), and Burger's equation (BE) which introduces an additional viscosity term. Furthermore, we use the full Navier-Stokes equations (NSE), which introduce a conservation of mass constraint. When combined with a deterministic solver and a suitable parameter step size, all these PDEs exhibit chaotic behaviour at small scales, and the medium to large scale characteristics of the solutions shift smoothly with increasing changes of the parameters $p_i$.

The noise amplifies the chaotic behaviour to larger scales and provides a controlled amount of perturbations for the data generation. This lets the network learn about the nature of the chaotic behaviour of PDEs without overwhelming it with data where patterns are not observable anymore.
The latter can easily happen when $\Delta_i$ or $v$ grow too large and produce essentially random outputs. Instead, we specifically target solutions that are difficult to evaluate in terms of a shallow metric. We heuristically select the smallest $v$ and a suitable $\Delta_i$ such that the ordering of several random output samples with respect to their $L^2$ difference drops below a correlation value of $0.8$. For the chosen PDEs, $v$ was small enough to avoid deterioration of the physical behaviour especially due to the diffusion terms, but different means of adjusting the difficulty may be necessary for other data.

\subsection{Training}
For training, the 2D scalar fields from the simulations were augmented with random flips, $90^{\circ}$ rotations, and cropping to obtain an input size of $224\times224$ every time they are used. Identical augmentations were applied to each field of one given sequence to ensure comparability. Afterwards, each input sequence is collectively normalized to the range $[0, 255]$. To allow for comparisons with image metrics and provide the possibility to compare color data and full velocity fields during inference, the metric uses three input channels. During training, the scalar fields are duplicated to each channel after augmentation.
Unless otherwise noted, networks were trained with a batch size of 1 for 40 epochs with the Adam optimizer using a learning rate of $10^{-5}$.
To evaluate the trained networks on validation and test inputs, only a bilinear resizing and the normalization step is applied.


\section{Correlation Loss Function}\label{sec: loss}

The central goal of our networks is to identify relative differences of input pairs produced via numerical simulations.
Thus, instead of employing a loss that forces the network to only infer given labels or distance values,
we train our networks to infer the ordering of a given sequence of simulation outputs $o_0, o_1, \dotsc, o_n$. 
We propose to use the Pearson correlation coefficient \citep[see][]{pearson1920}, which yields a value in $[-1,1]$ that measures the linear relationship between two distributions. A value of $1$ implies that a linear equation describes their relationship perfectly. We compute this coefficient for a full series of outputs such that the network can learn to extract features that arrange this data series in the correct ordering.
Each training sample of our network consists of every possible pair from the sequence $o_0, o_1, \dotsc, o_n$ and the corresponding ground truth distance distribution $\vc \in [0,1]^{0.5 (n+1) n}$ representing the parameter change from the data generation. For a distance prediction $\vd \in [0,\infty)^{0.5 (n+1) n}$ of our network for one sample, we compute the loss with:
\begin{equation}
    L(\vc, \vd) = \lambda_1 (\vc - \vd)^2 + \lambda_2 (1 - \frac{(\vc - \bar{\vc}) \cdot (\vd - \bar{\vd})}{\left\|\vc - \bar{\vc}\right\|_2 \left\|\vd - \bar{\vd}\right\|_2} )
    \label{eq: training loss}
\end{equation}

Here, the mean of a distance vector is denoted by $\bar{\vc}$ and $\bar{\vd}$ for ground truth and prediction, respectively. The first part of the loss is a regular MSE term, which minimizes the difference between predicted and actual distances. The second part is the Pearson correlation coefficient, which is inverted such that the optimization results in a maximization of the correlation.
As this formulation depends on the length of the input sequence, the two terms are scaled to adjust their relative influence with $\lambda_1$ and $\lambda_2$. For the training, we chose $n=10$ variations for each reference simulation. If $n$ should vary during training, the influence of both terms needs to be adjusted accordingly. We found that scaling both terms to a similar order of magnitude worked best in our experiments.

\begin{figure}[ht]
    \centering
    \includegraphics[width=0.48\textwidth]{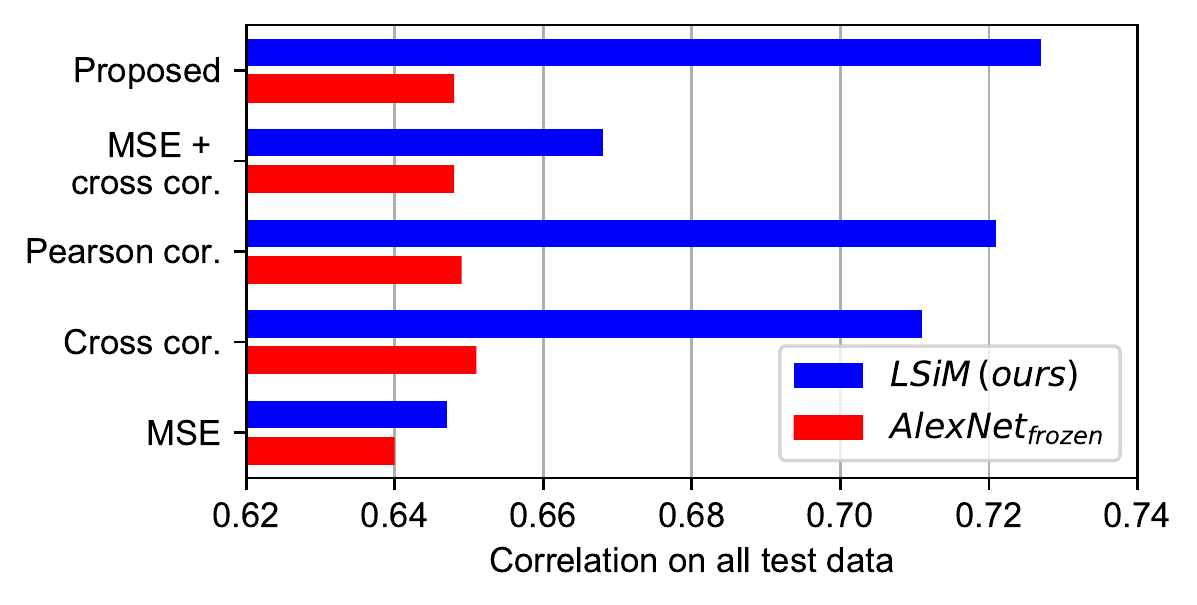}
    \vspace{-0.7cm}
    \caption{Performance comparison on our test data of the proposed approach (\textit{LSiM}) and a smaller model (\textit{AlexNet\textsubscript{frozen}}) for different loss functions on the y-axis.}
    \label{fig: correlation loss}
\end{figure}

In Fig.~\ref{fig: correlation loss}, we investigate how the proposed loss function compares to other commonly used loss formulations for our full network and a pre-trained network, where only aggregation weights are learned. The performance is measured via Spearman's rank correlation of predicted against ground truth distances on our combined test data sets. This is comparable to the \texttt{All} column in Tab.~\ref{table: results} and described in more detail in Section \ref{subsec: evaluation}. In addition to our full loss function, we consider a loss function that replaces the Pearson correlation with a simpler cross-correlation $(\vc \cdot \vd) \,/\, (\left\|\vc\right\|_2 \left\|\vd\right\|_2)$. We also include networks trained with only the MSE or only the correlation terms for each of the two variants.

A simple MSE loss yields the worst performance for both evaluated models. Using any correlation based loss function for the \textit{AlexNet\textsubscript{frozen}} metric (see Section \ref{subsec: evaluation}) improves the results, but there is no major difference due to the limited number of only 1152 trainable weights. For \textit{LSiM}, the proposed combination of MSE loss with the Pearson correlation performs better than using cross-correlation or only isolated Pearson correlation.
Interestingly, combining cross correlation with MSE yields worse results than cross correlation by itself. 
This is caused by the cross correlation term influencing absolute distance values, which potentially conflicts with the MSE term.
For our loss, the Pearson correlation only handles the relative ordering while the MSE deals with the absolute distances, leading to better inferred distances.


\section{Results} \label{sec: results}
In the following, we will discuss how the data generation approach was employed to create a large range of training and test data from different PDEs. Afterwards, the proposed metric is compared to other metrics, and its robustness is evaluated with several external data sets.

\begin{figure*}[bp]
    \centering
    \vspace{-0.2cm}
    \includegraphics[width=0.99\textwidth]{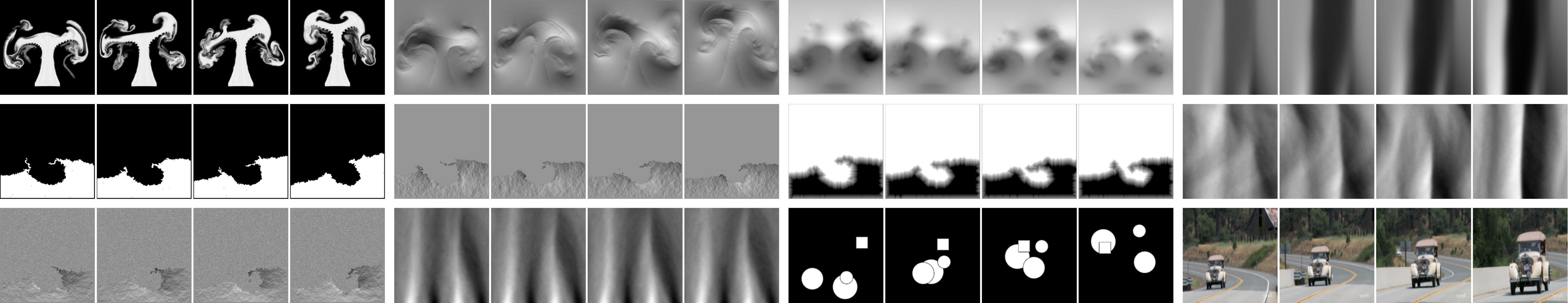}
    \vspace{-0.35cm}
    \caption{Samples from our data sets. For each subset the reference is on the left, and three variations in equal parameter steps follow. From left to right and top to bottom: \texttt{Smo} (density, velocity, and pressure), \texttt{Adv} (density), \texttt{Liq} (flags, velocity, and levelset), \texttt{Bur} (velocity), \texttt{LiqN} (velocity), \texttt{AdvD} (density), \texttt{Sha} and \texttt{Vid}.}
    \label{fig: data examples}
\end{figure*}

\subsection{Data Sets}
We created four training (\texttt{Smo}, \texttt{Liq}, \texttt{Adv} and \texttt{Bur}) and two test data sets (\texttt{LiqN} and \texttt{AdvD}) with ten parameter steps for each reference simulation. Based on two 2D NSE solvers, the smoke and liquid simulation training sets (\texttt{Smo} and \texttt{Liq}) add noise to the velocity field and feature varied initial conditions such as fluid position or obstacle properties, in addition to variations of buoyancy and gravity forces. 
The two other training sets (\texttt{Adv} and \texttt{Bur}) are based on 1D solvers for AD and BE, concatenated over time to form a 2D result. In both cases, noise was injected into the velocity field, and the varied parameters are changes to the field initialization and forcing functions. 

For the test data set, we substantially change the data distribution by injecting noise into the density instead of the velocity field for AD simulations to obtain the \texttt{AdvD} data set and by including background noise for the velocity field of a liquid simulation (\texttt{LiqN}). In addition, we employed three more test sets (\texttt{Sha}, \texttt{Vid}, and \texttt{TID}) created without PDE models to explore the generalization for data far from our training data setup.
We include a shape data set (\texttt{Sha}) that features multiple randomized moving rigid shapes, a video data set (\texttt{Vid}) consisting of frames from random video footage, and TID2013 \citep{ponomarenko2015} as a perceptual image data set (\texttt{TID}).
Below, we additionally list a combined correlation score (\texttt{All}) for all test sets apart from \texttt{TID}, which is excluded due to its different structure. Examples for each data set are shown in Fig.~\ref{fig: data examples} and generation details with further samples can be found in App.~\ref{append: data sets}.

\subsection{Performance Evaluation} \label{subsec: evaluation}
To evaluate the performance of a metric on a data set, we first compute the distances from each reference simulation to all corresponding variations. Then, the predicted and the ground truth distance distributions over all samples are combined and compared using Spearman's rank correlation coefficient \citep[see][]{spearman1904}. It is similar to the Pearson correlation, but instead it uses ranking variables, i.e., measures monotonic relationships of distributions.

\begin{table*}[tp]
    \vspace{-0.2cm}
    \caption{Performance comparison of existing metrics (top block), experimental designs (middle block), and variants of the proposed method (bottom block) on validation and test data sets measured in terms of Spearman's rank correlation coefficient of ground truth against predicted distances. \best{Bold+underlined} values show the best performing metric for each data set, \bestErr{bold} values are within a $0.01$ error margin of the best performing, and \bad{italic} values are $0.2$ or more below the best performing. On the right, a visualization of the combined test data results is shown for selected models.}
    \label{table: results}
    \centering
    \begin{tabular}[b]{l c c c c | c | c c c c c}
        \toprule
        \multirow{2}{*}[-1.3mm]{\bf Metric} & \multicolumn{4}{c |}{\bf Validation data sets} & \multicolumn{6}{c}{\bf Test data sets} \\
        \cmidrule(lr){2-5} \cmidrule(lr){6-11}
        & \texttt{Smo} & \texttt{Liq} & \texttt{Adv} & \texttt{Bur} & \texttt{TID} & \texttt{LiqN} & \texttt{AdvD} & \texttt{Sha} & \texttt{Vid} & \texttt{All}\\
        \cmidrule(lr){1-11}

        \it $L^2$                             & 0.66 & 0.80 & 0.74 & 0.62  & 0.82 & 0.73 & 0.57 & \bad{0.58} & 0.79 & 0.61 \\
        \it SSIM                              & 0.69 & 0.73 & 0.77 & 0.71  & 0.77 & \bad{0.26} & \best{0.69} & \bad{0.46} & 0.75 & \bad{0.53} \\
        \it LPIPS v0.1.                       & 0.63 & 0.68 & 0.68 & 0.72  & \best{0.86} & \bad{0.50} & 0.62 & 0.84 & \bestErr{0.83} & 0.66 \\
        \cmidrule(lr){2-11}

        \it AlexNet\textsubscript{random}     & 0.63 & 0.69 & 0.69 & 0.66  & 0.82 & 0.64 & 0.65 & \bad{0.67} & 0.81 & 0.65 \\
        \it AlexNet\textsubscript{frozen}     & 0.66 & 0.70 & 0.69 & 0.71  & \bestErr{0.85} & \bad{0.40} & 0.62 & 0.87 & \best{0.84} & 0.65 \\
        \it Optical flow                      & 0.62 & \bad{0.57} & \bad{0.36} & \bad{0.37}  & \bad{0.55} & \bad{0.49} & \bad{0.28} & \bad{0.61} & 0.75 & \bad{0.48} \\
        \it Non-Siamese                       & 0.77 & \best{0.85} & 0.78 & \bestErr{0.74}  & \bad{0.65} & \best{0.81} & 0.64 & \bad{0.25} & 0.80 & 0.60 \\
        \it Skip\textsubscript{from scratch}  & \best{0.79} & 0.83 & \best{0.80} & \bestErr{0.74}  & \bestErr{0.85} & 0.78 & 0.61 & 0.78 & \bestErr{0.83} & 0.71 \\
        \cmidrule(lr){2-11}
        
        \it LSiM\textsubscript{noiseless}     & 0.77 & 0.77 & 0.76 & 0.72  & \bestErr{0.85} & 0.62 & 0.58 & 0.86 & 0.82 & 0.68 \\
        \it LSiM\textsubscript{strong noise}  & 0.65 & \bad{0.65} & 0.67 & 0.69  & 0.84 & \bad{0.39} & 0.54 & \best{0.89} & 0.82 & 0.64 \\
        \it LSiM (ours)                       & \bestErr{0.78} & 0.82 & \bestErr{0.79} & \best{0.75}  & \bestErr{0.86} & 0.79 & 0.58 & \bestErr{0.88} & 0.81 & \best{0.73} \\
   
        \bottomrule
    \end{tabular}
    \hspace{-0.0cm}
    \includegraphics[width=0.187\textwidth]{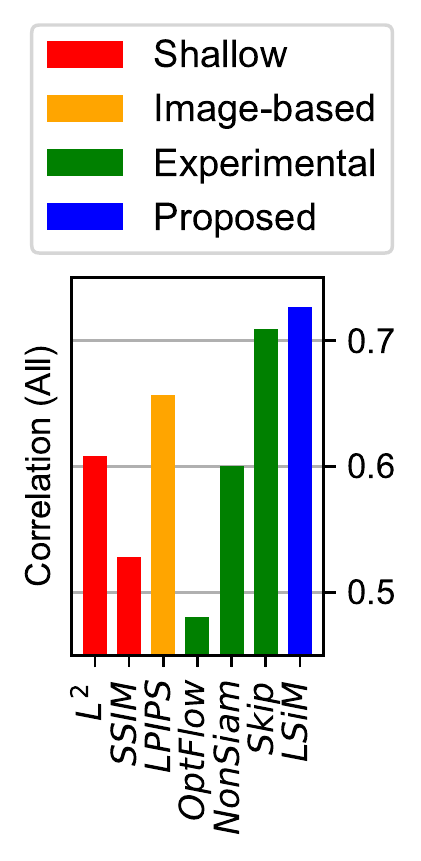}
\end{table*}

The top part of Tab.~\ref{table: results} shows the performance of the shallow metrics \textit{$L^2$} and \textit{SSIM} as well as the \textit{LPIPS} metric \citep{zhang2018} for all our data sets. The results clearly show that shallow metrics are not suitable to compare the samples in our data set and only rarely achieve good correlation values. The perceptual \textit{LPIPS} metric performs better in general and outperforms our method on the image data sets \texttt{Vid} and \texttt{TID}. This is not surprising as \textit{LPIPS} is specifically trained for such images. For most of the simulation data sets, however, it performs significantly worse than for the image content. The last row of Tab.~\ref{table: results} shows the results of our \textit{LSiM} model with a very good performance across all data sets and no negative outliers. Note that although it was not trained with any natural image content, it still performs well for the image test sets.

The middle block of Tab.~\ref{table: results} contains several interesting variants (more details can be found in App.~\ref{append: architectures}):
\textit{AlexNet\textsubscript{random}} and \textit{AlexNet\textsubscript{frozen}} are small models, where the base network is the original AlexNet with pre-trained weights. \textit{AlexNet\textsubscript{random}} contains purely random aggregation weights without training, whereas \textit{AlexNet\textsubscript{frozen}} only has trainable weights for the channel aggregation and therefore lacks the flexibility to fully adjust to the data distribution of the numerical simulations. The random model performs surprisingly well in general, pointing to powers of the underlying Siamese CNN architecture.

Recognizing that many PDEs include transport phenomena, we investigated optical flow \citep{horn1981} as a means to compute motion from field data. For the \textit{Optical flow} metric, we used FlowNet2 \citep{ilg2016} to bidirectionally compute the optical flow field between two inputs and aggregate it to a single distance value by summing all flow vector magnitudes. On the data set \texttt{Vid} that is similar to the training data of FlowNet2, it performs relatively well, but in most other cases it performs poorly. This shows that computing a simple warping from one input to the other is not enough for a stable metric although it seems like an intuitive solution. A more robust metric needs the knowledge of the underlying features and their changes to generalize better to new data. 

To evaluate whether a Siamese architecture is really beneficial, we used a \textit{Non-Siamese} architecture that directly predicts the distance from both stacked inputs. For this purpose, we employed a modified version of AlexNet that reduces the weights of the feature extractor by 50\% and of the remaining layers by 90\%. As expected, this metric works great on the validation data but has huge problems with generalization, especially on \texttt{TID} and \texttt{Sha}. In addition, even simple metric properties such as symmetry are no longer guaranteed because this architecture does not have the inherent constraints of the Siamese setup. 
Finally, we experimented with multiple fully trained base networks. As re-training existing feature extractors only provided small improvements, we used a custom base network with skip connections for the \textit{Skip\textsubscript{from scratch}} metric. Its results already come close to the proposed approach on most data sets.

The last block in Tab.~\ref{table: results} shows variants of the proposed approach trained with varied noise levels. This inherently changes the difficulty of the data. Hence, \textit{LSiM\textsubscript{noiseless}} was trained with relatively simple data without perturbations, whereas \textit{LSiM\textsubscript{strong noise}} was trained with strongly varying data. Both cases decrease the capabilities of the trained model on some of the validation and test sets. This indicates that the network needs to see a certain amount of variation at training time in order to become robust, but overly large changes hinder the learning of useful features (also see App.~\ref{append: noise}).

\subsection{Evaluation on Real-World Data} \label{subsec: real-world evaluation}

\begin{figure*}[tp]
    \centering
    \includegraphics[width=1.0\textwidth]{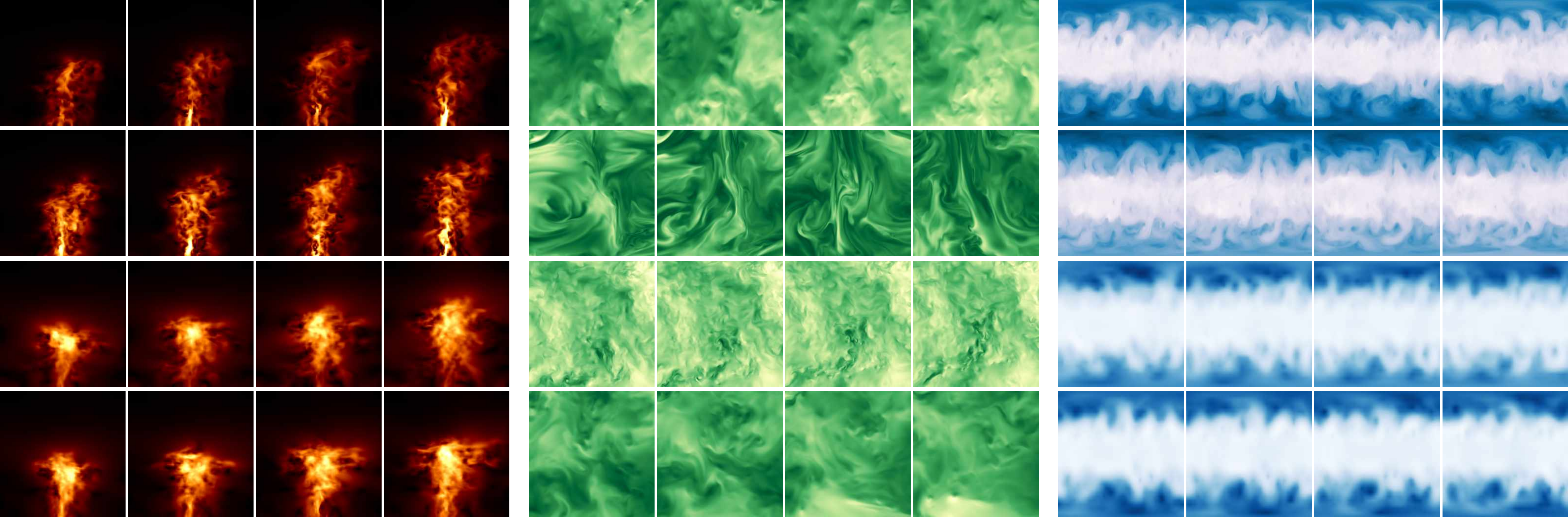}
    \vspace{-0.5cm}
    \caption{Examples from three real-world data repositories used for evaluation, visualized via color-mapping. Each block features four different sequences (rows) with frames in equal temporal or spatial intervals. Left: \textit{ScalarFlow} -- captured buoyant volumetric transport flows using the z-slice (top two) and z-mean (bottom two). Middle: \textit{JHTDB} -- four different turbulent DNS simulations. Right: \textit{WeatherBench} -- weather data consisting of temperature (top two) and geopotential (bottom two).} \label{fig: real world data}
\end{figure*}

To evaluate the generalizing capabilities of our trained metric, we turn to three representative and publicly available data sets of captured and simulated real-world phenomena, namely buoyant flows, turbulence, and weather. For the former, we make use of the \textit{ScalarFlow} data set \citep{eckert2019}, which consists of captured velocities of buoyant scalar transport flows.
Additionally, we include velocity data from the Johns Hopkins Turbulence Database (\textit{JHTDB}) \citep{perlman2007}, which represents direct numerical simulations of fully developed turbulence. As a third case, we use scalar temperature and geopotential fields from the \textit{WeatherBench} repository \citep{rasp2020}, which contains global climate data on a Cartesian latitude-longitude grid of the earth. Visualizations of this data via color-mapping the scalar fields or velocity magnitudes are shown in Fig.~\ref{fig: real world data}.

\begin{figure}[ht]
    \centering
    \vspace{-0.1cm}
    \includegraphics[width=0.48\textwidth]{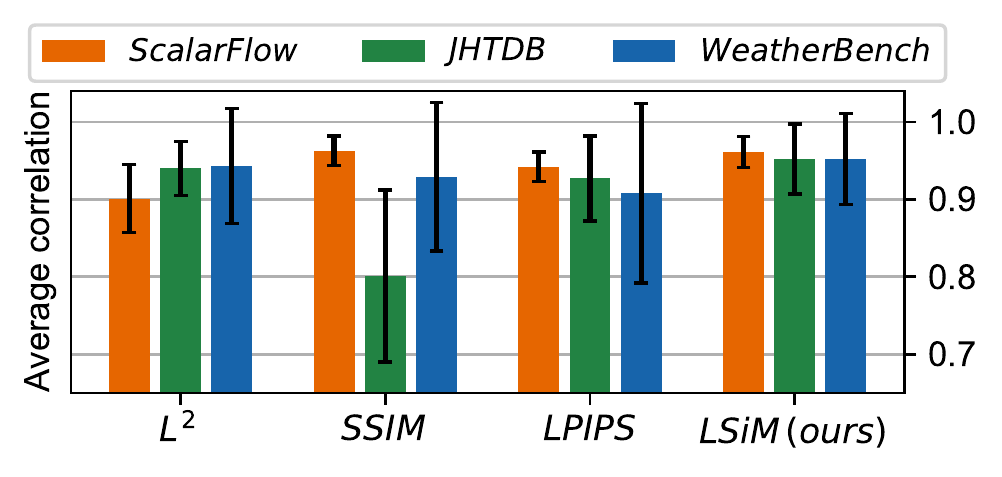}
    \vspace{-0.9cm}
    \caption{Spearman correlation values for multiple metrics on data from three repositories. Shown are mean and standard deviation over different temporal or spatial intervals used to create sequences.} \label{fig: real world evaluation}
\end{figure}

For the results in Fig.~\ref{fig: real world evaluation}, we extracted sequences of frames with fixed temporal and spatial intervals from each data set to obtain a ground truth ordering. Six different interval spacings for every data source are employed, and all velocity data is split by component. We then measure how well different metrics recover the original ordering in the presence of the complex changes of content, driven by the underlying physical processes. The \textit{LSiM} model outlined in previous sections was used for inference without further changes.

Every metric is separately evaluated (see Section \ref{subsec: evaluation}) for the six interval spacings with 180-240 sequences each. For \textit{ScalarFlow} and \textit{WeatherBench}, the data was additionally partitioned by z-slice or z-mean and temperature or geopotential respectively, leading to twelve evaluations. Fig.~\ref{fig: real world evaluation} shows the mean and standard deviation of the resulting correlation values.
Despite never being trained on any data from these data sets, \textit{LSiM} recovers the ordering of all three cases with consistently high accuracy. It yields averaged correlations of $0.96 \pm 0.02$, $0.95 \pm 0.05$, and $0.95 \pm 0.06$ for \textit{ScalarFlow}, \textit{JHTDB}, and \textit{WeatherBench}, respectively. The other metrics show lower means and higher uncertainty. Further details and results for the individual evaluations can be found in App.~\ref{append: real-world data}.


\section{Conclusion} \label{sec: conclusion}
We have presented the \textit{LSiM} metric to reliably and robustly compare outputs from numerical simulations. Our method significantly outperforms existing shallow metric functions and provides better results than other learned metrics. We demonstrated the usefulness of the correlation loss, showed the benefits of a controlled data generation environment, and highlighted the stability of the obtained metric for a range of real-world data sets. 

Our trained \textit{LSiM} metric has the potential to impact a wide range of fields, including the fast and reliable accuracy assessment of new simulation methods, robust optimizations of parameters for reconstructions of observations, and guiding generative models of physical systems.
Furthermore, it will be highly interesting to evaluate other loss functions, e.g., mutual information \citep{bachman2019} or contrastive predictive coding \citep{henaff2019}, and combinations with evaluations from perceptual studies \citep{um2019}. We also plan to evaluate our approach for an even larger set of PDEs as well as for 3D and 4D data sets.
Especially, turbulent flows are a highly relevant and interesting area for future work on learned evaluation metrics.

\section*{Acknowledgements}
This work was supported by the ERC Starting Grant \textit{realFlow} (StG-2015-637014). We would like to thank Stephan Rasp for preparing the \textit{WeatherBench} data and all reviewers for helping to improve this work.

\hbadness=99999 

{
\small
\bibliography{bibliography}
\bibliographystyle{icml2020}
}

\newpage

\appendix

\renewcommand\thesection{\Alph{section}}

\twocolumn[
\icmltitle{Appendix: Learning Similarity Metrics for Numerical Simulations}

\icmlkeywords{metric learning, CNNs, PDEs, numerical simulation, perceptual evaluation, physics simulation}
]

This supplemental document contains an analysis of the proposed metric design with respect to properties of metrics in general (App.~\ref{append: metric properties}) and details to the used network architectures (App.~\ref{append: architectures}). Afterwards, material that deals with the data sets is provided. It contains examples and failure cases for each of the data domains and analyzes the impact of the data difficulty (App.~\ref{append: noise} and \ref{append: data sets}). Next, the evaluation on real-world data is described in more detail (App.~\ref{append: real-world data}).
Finally, we explore additional metric evaluations (App.~\ref{append: additional evaluations}) and give an overview on the used notation (App.~\ref{append: notation}).

The source code for using the trained \textit{LSiM} metric and re-training the model from scratch are available at \url{https://github.com/tum-pbs/LSIM}. This includes the full data sets and the corresponding data generation scripts for the employed PDE solver.


\section{Discussion of Metric Properties} \label{append: metric properties}
To analyze if the proposed method qualifies as a \emph{metric}, it is split in two functions $m_1 : \sI \to \sL$ and $m_2 : \sL \times \sL \to [0, \infty)$, which operate on the input space $\sI$ and the latent space $\sL$. Through flattening elements from the input or latent space into vectors, $\sI \simeq \sR^a$ and $\sL \simeq \sR^b$ where $a$ and $b$ are the dimensions of the input data and all feature maps respectively, and both values have a similar order of magnitude. $m_1$ describes the non-linear function computed by the base network combined with the following normalization and returns a point in the latent space. $m_2$ uses two points in the latent space to compute a final distance value, thus it includes the latent space difference and the aggregation along the spatial, layer, and channel dimensions. With the Siamese network architecture, the resulting function for the entire approach is
\begin{equation*}
    m(\vx,\vy) \;=\; m_2( m_1(\vx), m_1(\vy) ).
\end{equation*}
The identity of indiscernibles mainly depends on $m_1$ because, even if $m_2$ itself guarantees this property, $m_1$ could still be non-injective, which means it can map different inputs to the same point in latent space $\tilde{\vx} = \tilde{\vy}$ for $\vx \neq \vy$. Due to the complicated nature of $m_1$, it is difficult to make accurate predictions about the injectivity of $m_1$. Each base network layer of $m_1$ recursively processes the result of the preceding layer with various feature extracting operations. Here, the intuition is that significant changes in the input should produce different feature map results in one or more layers of the network. As very small changes in the input lead to zero valued distances predicted by the CNN (i.e., an identical latent space for different inputs), $m_1$ is in practice not injective. In an additional experiment, the proposed architecture was evaluated on about 3500 random inputs from all our data sets, where the CNN received one unchanged and one slightly modified input. The modification consisted of multiple pixel adjustments by one bit (on 8-bit color images) in random positions and channels. When adjusting only a single pixel in the $224\times224$ input, the CNN predicts a zero valued distance on about 23\% of the inputs, but we never observed an input where seven or more changed pixels resulted in a distance of zero in all experiments. 

In this context, the problem of numerical errors is important because even two slightly different latent space representations could lead to a result that seems to be zero if the difference vanishes in the aggregation operations or is smaller than the floating point precision. On the other hand, an automated analysis to find points that have a different input but an identical latent space image is a challenging problem and left as future work.

The evaluation of the base network and the normalization is deterministic, and hence $\forall \vx : m_1(\vx) = m_1(\vx)$ holds. Furthermore, we know that $m(\vx,\vx) = 0$ if $m_2$ guarantees that $\forall m_1(\vx) : m_2( m_1(\vx), m_1(\vx) ) = 0$. Thus, the remaining properties, i.e., non-negativity, symmetry, and the triangle inequality, only depend on $m_2$ since for them the original inputs are not relevant, but their respective images in the latent space.
The resulting structure with a relaxed identity of indiscernibles is called a \emph{pseudometric}, where $\forall \tilde{\vx},\tilde{\vy},\tilde{\vz} \in \sL$:
\begin{align}
    m_2(\tilde{\vx},\tilde{\vy}) \; &\geq \; 0                \label{eq: relaxed NonNeg}\\
    m_2(\tilde{\vx},\tilde{\vy}) \; &= \; m_2(\tilde{\vy},\tilde{\vx})              \label{eq: relaxed Sym}\\
    m_2(\tilde{\vx},\tilde{\vy}) \; &\leq \; m_2(\tilde{\vx},\tilde{\vz}) + m_2(\tilde{\vz},\tilde{\vy}) \label{eq: relaxed TriIneq}\\
    m_2(\tilde{\vx},\tilde{\vx}) \; &= \; 0                   \label{eq: relaxed IoI}
\end{align}
Notice that $m_2$ has to fulfill these properties with respect to the latent space but not the input space. If $m_2$ is carefully constructed, the metric properties still apply, independently of the actual design of the base network or the feature map normalization.

\begin{figure*}[hbp]
    \centering
    \includegraphics[width=1.0\textwidth]{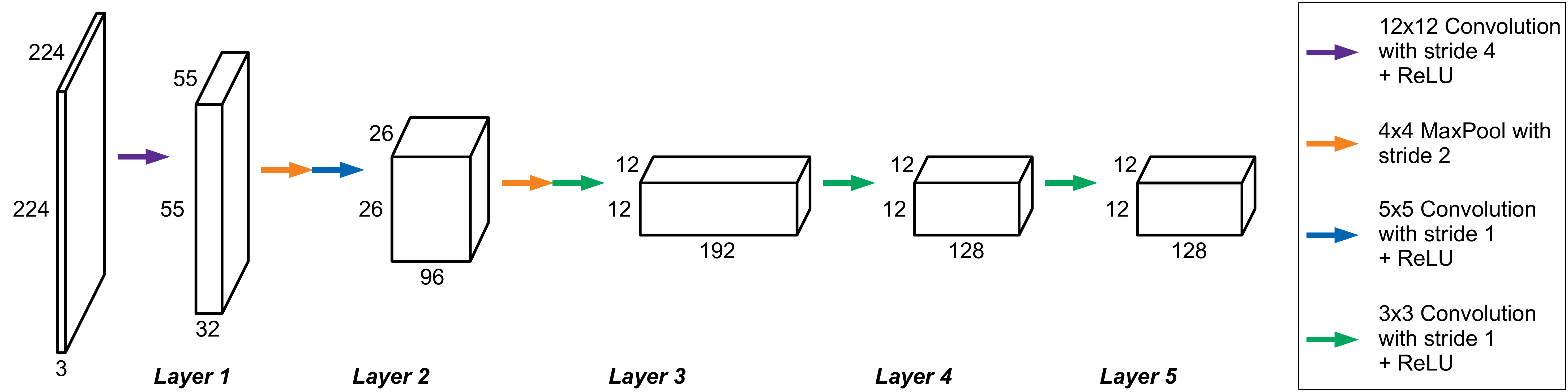}
    \vspace{-0.5cm}
    \caption{Proposed base network architecture consisting of five layers with up to 192 feature maps that are decreasing in spatial size. It is similar to the feature extractor from AlexNet as identical spatial dimensions for the feature maps are used, but it reduces the number of feature maps for each layer by 50\% to have fewer weights.}
    \label{fig: base network}
\end{figure*}

A first observation concerning $m_2$ is that if all aggregations were sum operations and the element-wise latent space difference was the absolute value of a difference operation, $m_2$ would be equivalent to computing the $L^1$ norm of the difference vector in latent space:
\begin{equation*}
    m_2^{sum}( \tilde{\vx},\tilde{\vy} ) \;=\; \sum_{i=1}^{b} \lvert \tilde{\vx}_i - \tilde{\vy}_i \rvert.
\end{equation*}
Similarly, adding a square operation to the element-wise distance in the latent space and computing the square root at the very end leads to the $L^2$ norm of the latent space difference vector. In the same way, it is possible to use any $L^p$ norm with the corresponding operations:
\begin{equation*} 
    m_2^{sum}( \tilde{\vx},\tilde{\vy} ) \;=\; \left( \sum_{i=1}^{b} \lvert \tilde{\vx}_i - \tilde{\vy}_i \rvert^p \right) ^ \frac{1}{p}.
\end{equation*}
In both cases, this forms the metric induced by the corresponding norm, which by definition has all desired properties (\ref{eq: relaxed NonNeg}), (\ref{eq: relaxed Sym}), (\ref{eq: relaxed TriIneq}), and (\ref{eq: relaxed IoI}). If we change all aggregation methods to a weighted average operation, each term in the sum is multiplied by a weight $w_i$. This is even possible with learned weights, as they are constant at evaluation time if they are clamped to be positive as described above. Now, $w_i$ can be attributed to both inputs by distributivity, meaning each input is element-wise multiplied with a constant vector before applying the metric, which leaves the metric properties untouched. The reason is that it is possible to define new vectors in the same space, equal to the scaled inputs. This renaming trivially provides the correct properties:
\begin{align*}
    &m_2^{weighted}( \tilde{\vx},\tilde{\vy} ) \;=\; \sum_{i=1}^{b} w_i \lvert \tilde{\vx}_i - \tilde{\vy}_i \rvert, \\ \;&\overset{w_i>0}{=}\; \sum_{i=1}^{b} \lvert w_i \tilde{\vx}_i - w_i \tilde{\vy}_i \rvert.
\end{align*}
Accordingly, doing the same with the $L^p$ norm idea is possible, and each $w_i$ just needs a suitable adjustment before distributivity can be applied, keeping the metric properties once again:
\begin{align*} 
    &m_2^{weighted}( \tilde{\vx},\tilde{\vy} ) \;=\; \left( \sum_{i=1}^{b} w_i \lvert \tilde{\vx}_i - \tilde{\vy}_i \rvert^p \right) ^ \frac{1}{p} \\
\end{align*}
\begin{align*}
    &=\; \left( \sum_{i=1}^{b} w_i \lvert \tilde{\vx}_i - \tilde{\vy}_i \rvert \; \lvert \tilde{\vx}_i - \tilde{\vy}_i \rvert \; \ldots \lvert \tilde{\vx}_i - \tilde{\vy}_i \rvert \right) ^ \frac{1}{p} \\
    &=\; \left( \sum_{i=1}^{b} w_i^\frac{1}{p} \lvert \tilde{\vx}_i - \tilde{\vy}_i \rvert \; w_i^\frac{1}{p} \lvert \tilde{\vx}_i - \tilde{\vy}_i \rvert \; \ldots w_i^\frac{1}{p} \lvert \tilde{\vx}_i - \tilde{\vy}_i \rvert \right) ^ \frac{1}{p}, \\
    &\overset{w_i>0}{=} \left( \sum_{i=1}^{b} \lvert w_i^\frac{1}{p} \tilde{\vx}_i - w_i^\frac{1}{p} \tilde{\vy}_i \rvert^p \right) ^ \frac{1}{p}.
\end{align*}
With these weighted terms for $m_2$, it is possible to describe all used aggregations and latent space difference methods. The proposed method deals with multiple higher order tensors instead of a single vector. Thus, the weights $w_i$ additionally depend on constants such as the direction of the aggregations and their position in the latent space tensors. But it is easy to see that mapping a higher order tensor to a vector and keeping track of additional constants still retains all properties in the same way. As a result, the described architecture by design yields a pseudometric that is suitable for comparing simulation data in a way that corresponds to our intuitive understanding of distances.


\section{Architectures} \label{append: architectures}
The following sections provide details regarding the architecture of the base network and some experimental design.

\subsection{Base Network Design} \label{append: base network}
Fig.~\ref{fig: base network} shows the architecture of the base network for the \textit{LSiM} metric. Its purpose is to extract features from both inputs of the Siamese architecture that are useful for the further processing steps. To maximise the usefulness and to avoid feature maps that show overly similar features, the chosen kernel size and stride of the convolutions are important. Starting with larger kernels and strides means the network has a big receptive field and can consider simple, low-level features in large regions of the input. For the two following layers, the large strides are replaced by additional MaxPool operations that serve a similar purpose and reduce the spatial size of the feature maps.

For the three final layers, only small convolution kernels and strides are used, but the number of channels is significantly larger than before. These deep features maps typically contain high-level structures, which are most important to distinguish complex changes in the inputs. Keeping the number of trainable weights as low as possible was an important consideration for this design to prevent overfitting to certain simulations types and increase generality. We explored a weight range by using the same architecture and only scaling the number of feature maps in each layer. The final design shown in Fig.~\ref{fig: base network} with about 0.62 million weights worked best for our experiments.

\begin{figure*}[ht]
    \centering
    \includegraphics[width=1.0\textwidth]{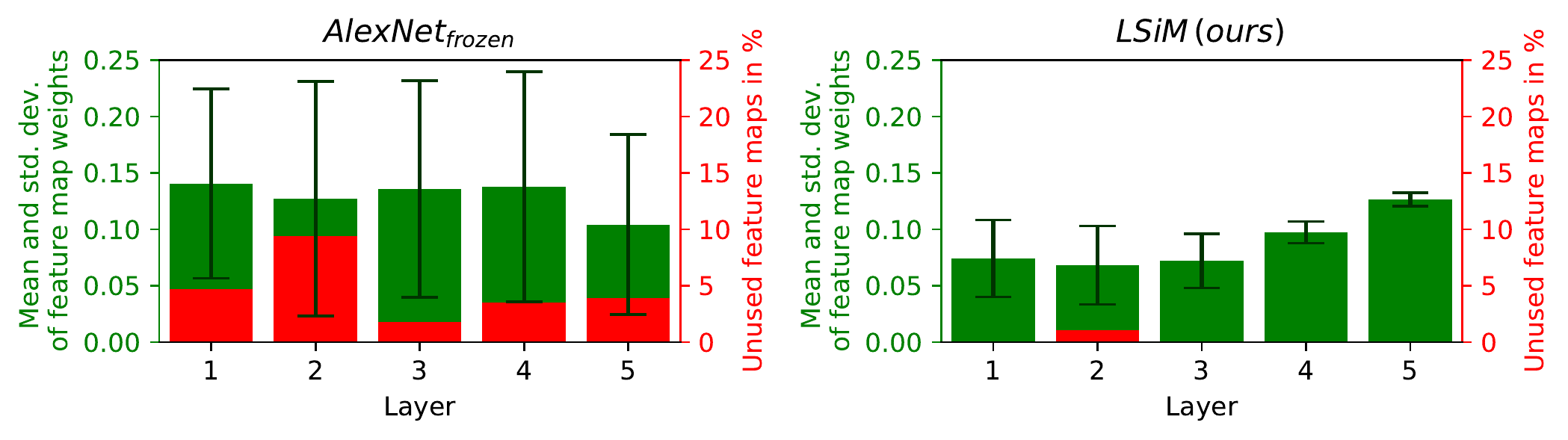}
    \vspace{-0.9cm}
    \caption{Analysis of the distributions of learned feature map aggregation weights across the base network layers. Displayed is a base network with pre-trained weights (left) in comparison to our method for fully training the base network (right). Note that the percentage of unused feature maps for most layers of our base network is 0\%.}
    \label{fig: feature map weights}
\end{figure*}

In the following, we analyze the contributions of the per-layer features of two different metric networks to highlight differences in terms of how the features are utilized for the distance estimation task.
In Fig.~\ref{fig: feature map weights}, our \textit{LSiM} network yields a significantly smaller standard deviation in the learned weights that aggregate feature maps of five layers, compared to a pre-trained base network. This means, all feature maps contribute to establishing the distances similarly, and the aggregation just fine-tunes the relative importance of each feature. In addition, almost all features receive a weight greater than zero, and as a result, more features are contributing to the final distance value.

Employing a fixed pre-trained feature extractor, on the other hand, shows a very different picture:
Although the mean across the different network layers is similar, the contributions of different features vary strongly, which is visible in the standard deviation being significantly larger. 
Furthermore, 2-10\% of the feature maps in each layer receive a weight of zero and 
hence were deemed not useful at all for establishing the distances.
This illustrates the usefulness of a targeted network in which all features contribute to the distance inference.

\subsection{Feature Map Normalization} \label{append: feature map norm}
\begin{figure}[ht]
    \centering
    \vspace{-0.2cm}
    \includegraphics[width=0.47\textwidth]{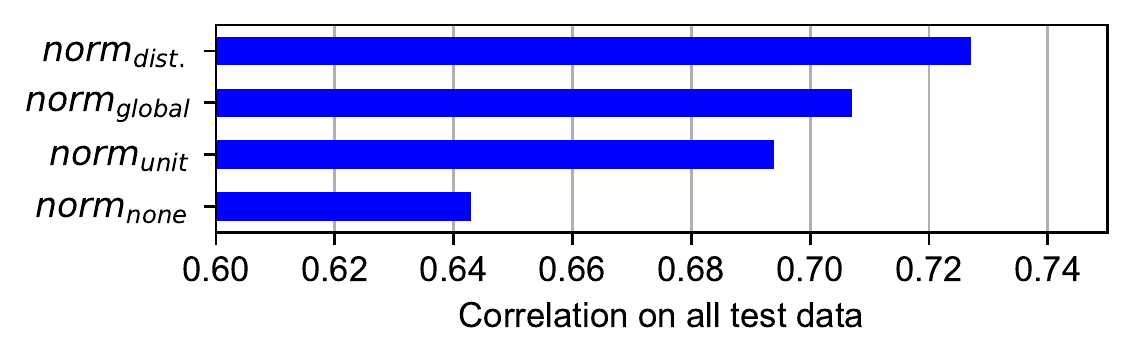}
    \vspace{-0.5cm}
    \caption{Performance on our test data for different feature map normalization approaches.}
    \label{fig: normalization}
    \vspace{-0.2cm}
\end{figure}

In the following, we analyze how the different feature map normalizations discussed in Section~\ref{subsec: feature map norm} of the main paper affect the performance of our metric. We compare using no normalization $norm_{none}(\tG) = \tG,$ the unit length normalization via division by the norm of a feature vector $norm_{unit}(\tG) = \tG \, / \left\| \tG \right\|_2$ proposed by \citeauthor{zhang2018}, a global unit length normalization $norm_{global}(\tG) = \tG \, / \, \text{max} \left(  \left\| \tG_0 \right\|_2,  \left\| \tG_1 \right\|_2, \dotsc  \right)$ that considers the norm of all feature vectors in the entire training set, and the proposed normalization to a scaled chi distribution
\begin{equation*}
    \text{norm}_{\text{dist}}(\tG) \;=\; \frac{1}{\sqrt{g_c - 1}} \, \frac{\tG - \text{mean} \left(  \tG_0, \tG_1, \dotsc  \right)}
    {\text{std} \left(  \tG_0, \tG_1, \dotsc  \right)}.
\end{equation*}
Fig.~\ref{fig: normalization} shows a comparison of these normalization methods on the combined test data. Using no normalization is significantly detrimental to the performance of the metric as succeeding operations cannot reliably compare the features. A unit length normalization of a single sample is already a major improvement since following operations now have a predictable range of values to work with. This corresponds to a cosine distance, which only measures angles of the feature vectors and entirely neglects their length.

Using the maximum norm across all training samples (computed in a pre-processing step and fixed for training) introduces additional information as the network can now compare magnitudes as well. However, this comparison is not stable as the maximum norm can be an outlier with respect to the typical content of the corresponding feature. The proposed normalization forms a chi distribution by individually transforming each component of the feature vector to a standard normal distribution. Afterwards, scaling with the inverse mode of the chi distribution leads to a consistent average magnitude close to one. It results in the best performing metric since both length and angle of the feature vectors can be reliably compared by the following operations.

\subsection{Recursive ``Meta-Metric''} \label{append: meta metric}
Since comparing the feature maps is a central operation of the proposed metric calculations, we experimented with replacing it with an existing CNN-based metric. In theory, this would allow for a recursive, arbitrarily deep network that repeatedly invokes itself: first, the extracted representations of inputs are used and then the representations extracted from the previous representations, etc. In practice, however, using more than one recursion step is currently not feasible due to increasing computational requirements in addition to vanishing gradients.

\begin{figure*}[bp]
    \centering
    \includegraphics[width=0.99\textwidth]{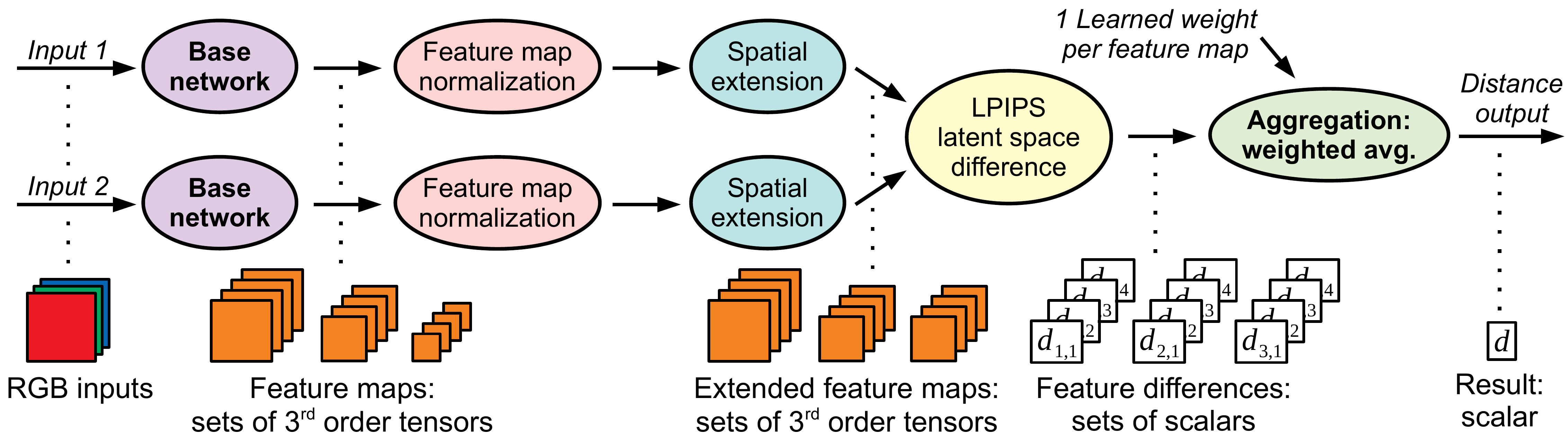}
    \caption{Adjusted distance computation for a \textit{LPIPS}-based latent space difference. To provide sufficiently large inputs for \textit{LPIPS}, small feature maps are spatially enlarged with nearest neighbor interpolation. In addition, \textit{LPIPS} creates scalar instead of spatial differences leading to a simplified aggregation.}
    \label{fig: distance latent lpips}
\end{figure*}

Fig.~\ref{fig: distance latent lpips} shows how our computation method can be modified for a CNN-based latent space difference, instead of an element-wise operation. Here we employ \textit{LPIPS} \citep{zhang2018}. There are two main differences compared to proposed method. First, the \textit{LPIPS} latent space difference creates single distance values for a pair of feature maps instead of a spatial feature difference. As a result, the following aggregation is a single learned average operation and spatial or layer aggregations are no longer necessary. We also performed experiments with a spatial \textit{LPIPS} version here, but due to memory limitations, these were not successful. Second, the convolution operations in \textit{LPIPS}
have a lower limit for spatial resolution, and some feature maps of our base network are quite small (see Fig.~\ref{fig: base network}). Hence, we up-scale the feature maps below the required spatial size of $32\times32$ using nearest neighbor interpolation.

On our combined test data, such a metric with a fully trained base network achieves a performance comparable to \textit{AlexNet\textsubscript{random}} or \textit{AlexNet\textsubscript{frozen}}.

\subsection{Optical Flow Metric}
\begin{figure*}[ht]
    \centering
    \includegraphics[width=0.75\textwidth]{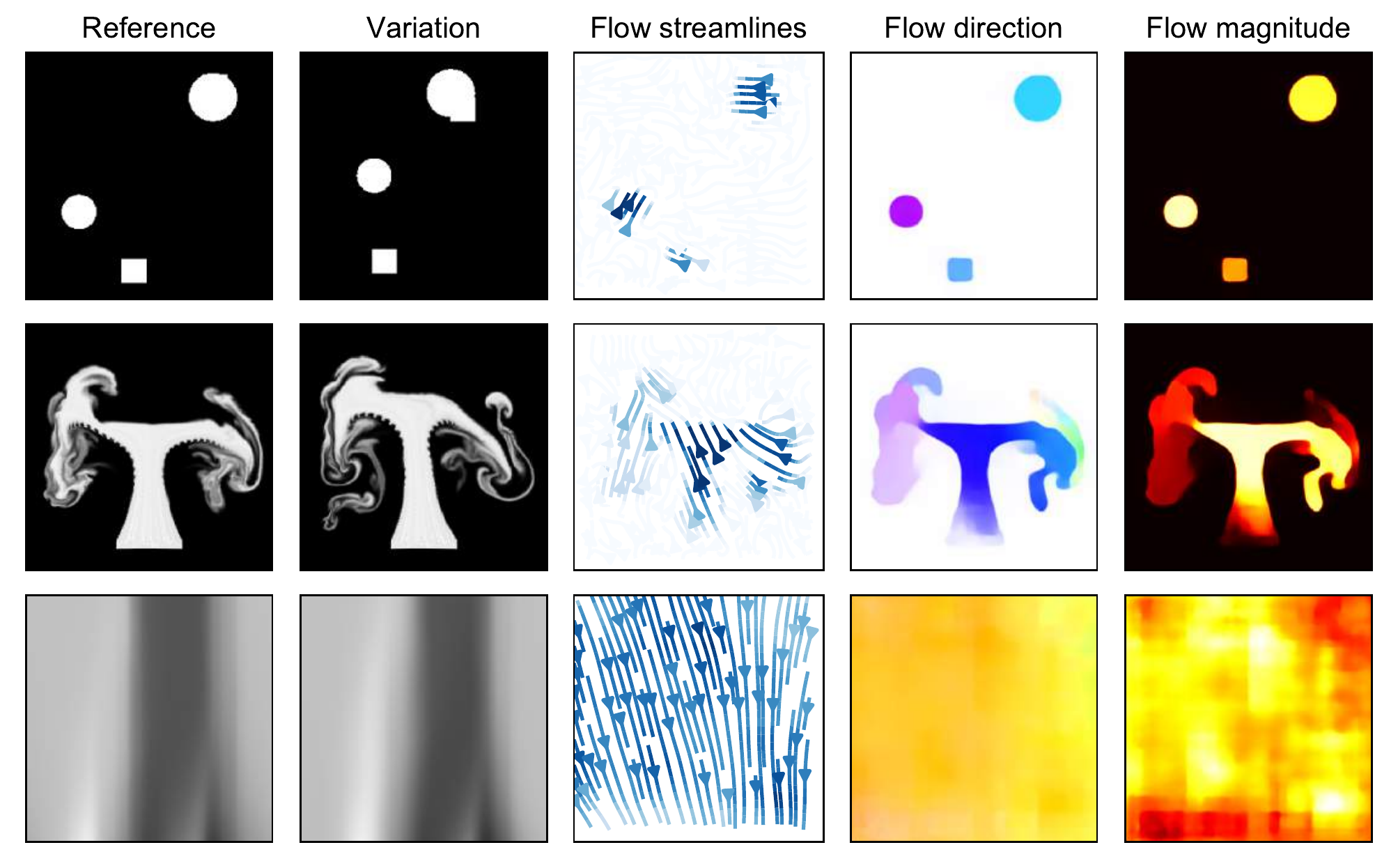}
    \vspace{-0.5cm}
    \caption{Outputs from FlowNet2 on data examples. The flow streamlines are sparse visualization of the resulting flow field and indicate the direction of the flow by their orientation and its magnitude by their color (darker being larger). The two visualizations on the right show the dense flow field and are color-coded to show the flow direction (blue/yellow: vertical, green/red: horizontal) and the flow magnitude (brighter being larger).}
    \label{fig: optical flow}
\end{figure*}
In the following, we describe our approach to compute a metric via optical flow (OF). For an efficient OF evaluation, we employed a pre-trained network \citep{ilg2016}.
From an OF network $f : \sI \times \sI \to \sR^{i_{max} \times j_{max} \times 2}$ with two input data fields $\vx,\vy \in I$, we get the flow vector field $f^{\vx\vy}(i,j) \,=\, ( f_1^{\vx\vy}(i,j),\, f_2^{\vx\vy}(i,j) )^T$, where $i$ and $j$ denote the locations, and $f_1$ and $f_2$ denote the components of the flow vectors. In addition, we have a second flow field $f^{\vy\vx}(i,j)$ computed from the reversed input ordering. We can now define a function $m : \sI \times \sI \to [0, \infty)$:
\begin{align*}
    m(\vx,\vy) \;=\; \sum_{i=0}^{i_{max}} \sum_{j=0}^{j_{max}} &\sqrt{(f^{\vx\vy}_1(i,j))^2 + (f^{\vx\vy}_2(i,j))^2} \\
    \,+\, &\sqrt{(f^{\vy\vx}_1(i,j))^2 + (f^{\vy\vx}_2(i,j))^2}.
\end{align*}
Intuitively, this function computes the sum over the magnitudes of all flow vectors in both vector fields. With this definition, it is obvious that $m(\vx,\vy)$ fulfills the metric properties of non-negativity and symmetry (see Eq.~(\ref{eq: relaxed NonNeg}) and (\ref{eq: relaxed Sym})). Under the assumption that identical inputs create a zero flow field, a relaxed identity of indiscernibles holds as well (see Eq.~(\ref{eq: relaxed IoI})). Compared to the proposed approach, there is no guarantee for the triangle inequality though, thus $m(\vx,\vy)$ only qualifies as a pseudo-semimetric.

Fig.~\ref{fig: optical flow} shows flow visualizations on data examples produced by FlowNet2. The metric works relatively well for inputs that are similar to the training data from FlowNet2 such as the shape data example in the top row. For data that provides some outline, e.g., the smoke simulation example in the middle row or also liquid data, the metric does not work as well but still provides a reasonable flow field. However, for full spatial examples such as the Burger's or Advection-Diffusion cases (see bottom row), the network is no longer able to produce meaningful flow fields. The results are often a very uniform flow with similar magnitude and direction.

\subsection{Non-Siamese Architecture}
\begin{figure*}[ht]
    \centering
    \includegraphics[width=0.99\textwidth]{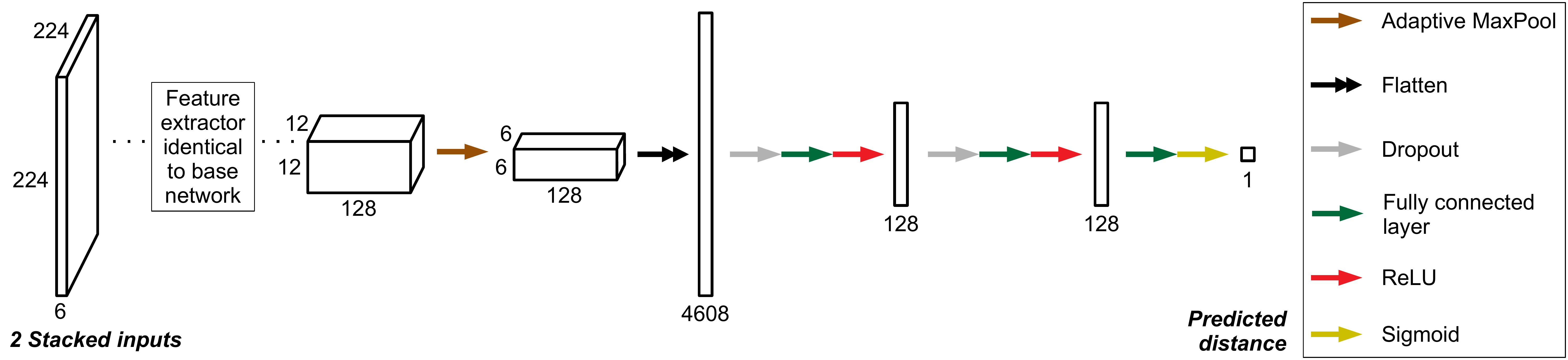}
    \caption{Non-Siamese network architecture with the same feature extractor used in Fig.~\ref{fig: base network}. It uses both stacked inputs and directly predicts the final distance value from the last set of feature maps with several fully connected layers.}
    \label{fig: non-siamese network}
\end{figure*}

To compute a metric without the Siamese architecture outlined above, we use a network structure with a single output as shown in 
Fig.~\ref{fig: non-siamese network}. Thus, instead of having two identically feature extractors and combining the feature maps, here the distance is directly predicted from the stacked inputs with a single network with about 1.24 million weights. After using the same feature extractor as described in Section~\ref{append: base network}, the final set of feature maps is spatially reduced with an adaptive MaxPool operation.
Next, the result is flattened, and three consecutive fully connected layers process the data to form the final prediction. Here, the last activation function is a sigmoid instead of ReLU. The reason is that a ReLU would clamp every negative intermediate value to a zero distance, while a sigmoid compresses the intermediate value to a small distance that is more meaningful than directly clamping it.
 
In terms of metric properties, this architecture only provides non-negativity (see Eq.~(\ref{eq: relaxed NonNeg})) due to the final sigmoid function. All other properties cannot be guaranteed without further constraints. This is the main disadvantage of a non-Siamese network. These issues could be alleviated with specialized training data or by manually adding constraints to the model, e.g., to have some amount of symmetry (see Eq.~(\ref{eq: relaxed Sym})) and at least a weakened identity of indiscernibles (see Eq.~(\ref{eq: relaxed IoI})). However, compared to a Siamese network that guarantees them by design, these extensions are clearly sub-optimal. As a result of the missing properties, this network has significant problems with generalization. While it performs well on the training data, the performance noticeably deteriorates for several of the test data sets.

\subsection{Skip Connections in Base Network}
\begin{figure*}[tp]
    \centering
    \includegraphics[width=0.9\textwidth]{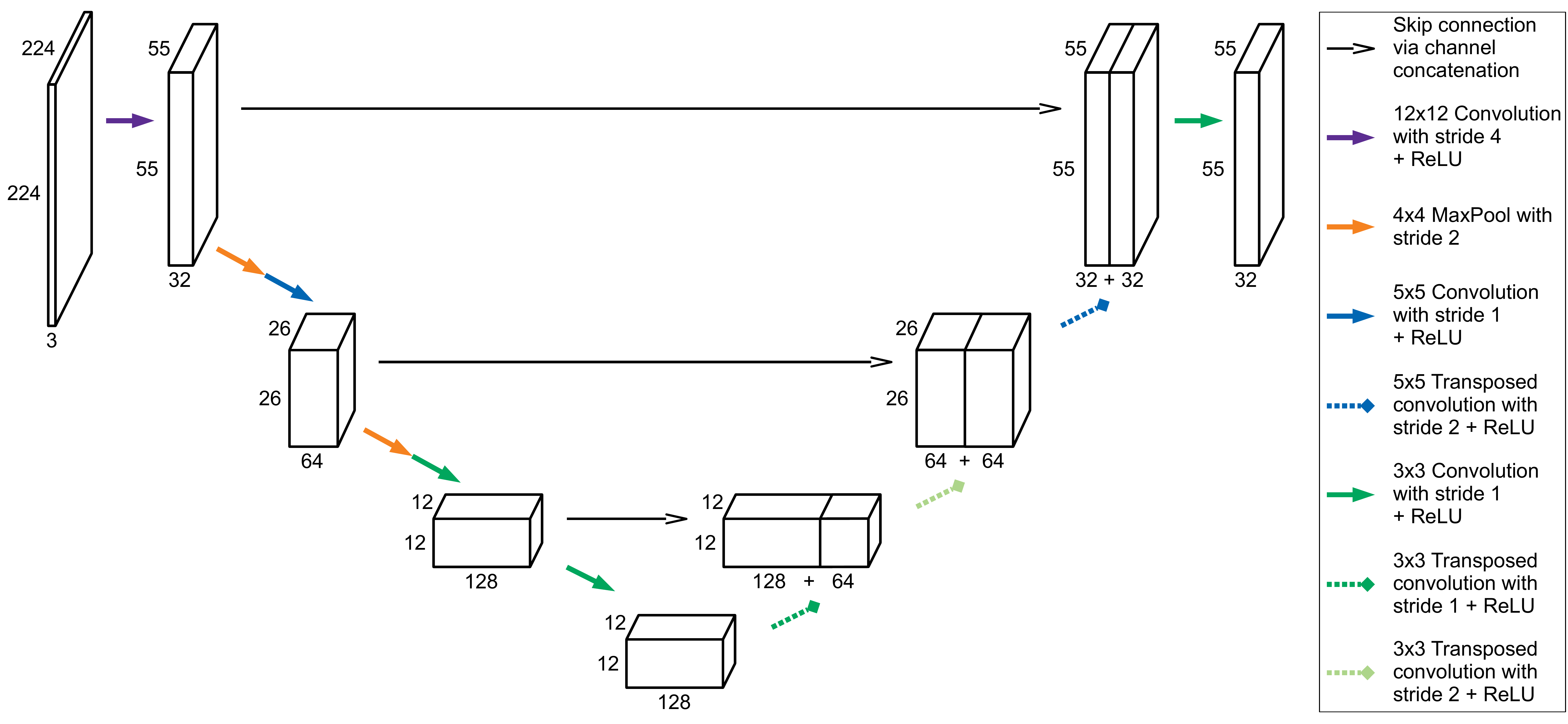}
    \caption{Network architecture with skip connections for better information transport between feature maps. Transposed convolutions are used to upscale the feature maps in the second half of the network to match the spatial size of earlier layers for the skip connections.}
    \label{fig: skip network}
\end{figure*}

As explained above, our base network primarily serves as a feature extractor to produce activations that are employed to evaluate a learned metric.In many state-of-the-art methods, networks with skip connections are employed \citep{ronneberger2015,he2016,huang2017}, as experiments have shown that these connections help to preserve information from the inputs. In our case, the classification ``output'' of a network such as the AlexNet plays no actual role. Rather, the features extracted along the way are crucial. Hence, skip connections should not improve the inference task for our metrics.

To verify that this is the case, we have included tests with a base network (see Fig.~\ref{fig: skip network}) similar to the popular UNet architecture \citep{ronneberger2015}. For our experiments, we kept the early layers closely in line with the feature extractors that worked well for the base network (see Section~\ref{append: base network}). Only the layers in the decoder part have an increased spatial feature map size to accommodate the skip connections. As expected, this network can be used to compute reliable metrics for the input data without negatively affecting the performance. However, as expected, the improvements of skip connections for regular inference tasks do not translate into improvements for the metric calculations.


\section{Impact of Data Difficulty} \label{append: noise}
\begin{figure}[ht]
    \centering
    \includegraphics[width=0.42\textwidth]{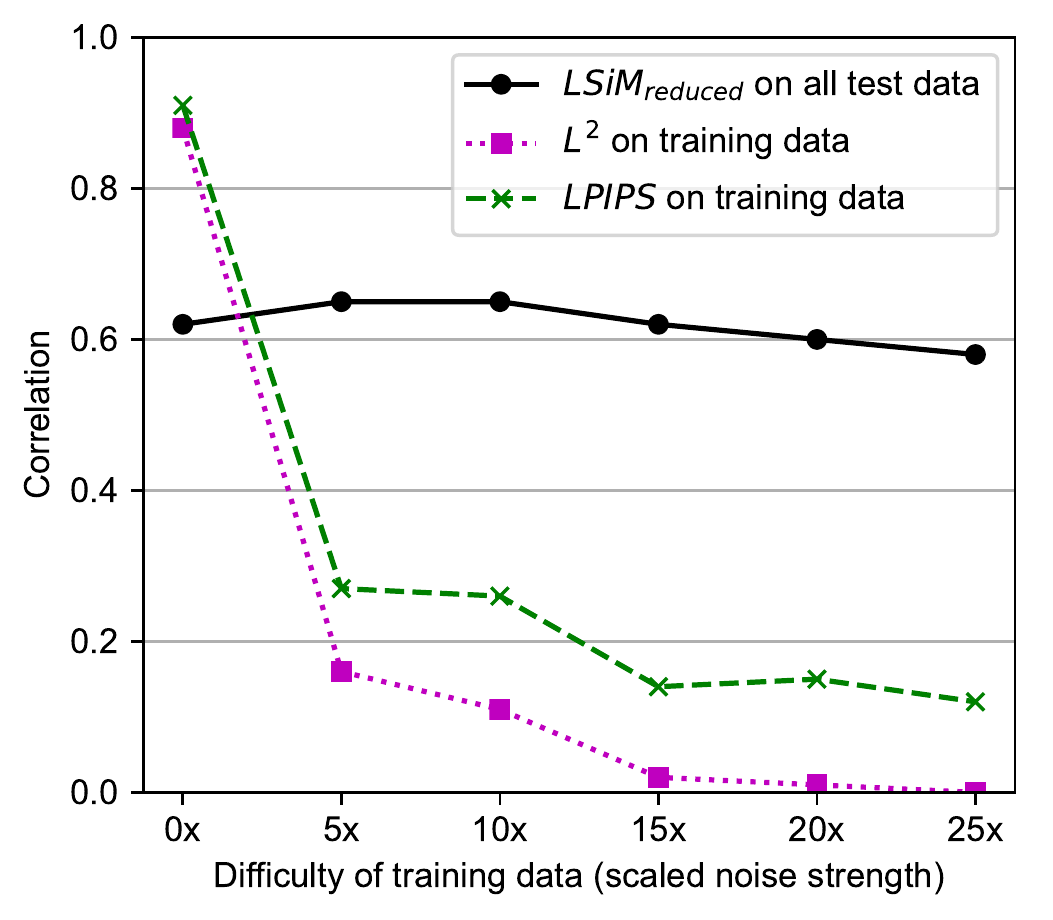}
    \vspace{-0.4cm}
    \caption{Impact of increasing data difficulty for a reduced training data set. 
    Evaluations on training data for $L^2$ and \textit{LPIPS}, and the test performance of models trained with the different reduced data sets (\textit{LSiM\textsubscript{reduced}}) are shown.}
    \label{fig: data difficulty}
\end{figure}
We shed more light on the aspect of noise levels and data difficulty via six reduced data sets that consist of a smaller amount of Smoke and Advection-Diffusion data with differently scaled noise strength values. Results are shown in Fig.~\ref{fig: data difficulty}.
Increasing the noise level creates more difficult data as shown by the dotted and dashed plots representing the performance of the \textit{$L^2$} and the \textit{LPIPS} metric on each data set. Both roughly follow an exponentially decreasing function. Each point on the solid line plot is the test result of a reduced \textit{LSiM} model trained on the data set with the corresponding noise level. Apart from the data, the entire training setup was identical. This shows that the training process is very robust to the noise, as the result on the test data only slowly decreases for very high noise levels. Furthermore, small amounts of noise improve the generalization compared to the model that was trained without any noise. This is somewhat expected, as a model that never saw noisy data during training cannot learn to extract features which are robust with respect to noise.


\section{Data Set Details} \label{append: data sets}
In the following sections, the generation of each used data set is described. For each figure showing data samples (consisting of a reference simulation and several variants with a single changing initial parameter), the leftmost image is the reference and the images to the right show the variants in order of increasing parameter change. For the figures~\ref{fig: smoke example}, \ref{fig: liquid example}, \ref{fig: advDiff example}, and \ref{fig: burgersEq example}, the first subfigure (a) demonstrates that medium and large scale characteristics behave very non-chaotic for simulations without any added noise. They are only included for illustrative purposes and are not used for training. The second and third subfigure (b) and (c) in each case show the training data of \textit{LSiM}, where the large majority of data falls into the category (b) of normal samples that follow the generation ordering, even with more varying behaviour. Category (c) is a small fraction of the training data, and the shown examples are specifically picked to show how the chaotic behaviour can sometimes override the ordering intended by the data generation in the worst case. Occasionally, category (d) is included to show how normal data samples from the test set differ from the training data.

\subsection{Navier-Stokes Equations}
These equations describe the general behaviour of fluids with respect to advection, viscosity, pressure, and mass conservation. Eq.~(\ref{eq: navier stokes momentum}) defines the conservation of momentum, and Eq.~(\ref{eq: navier stokes mass}) constraints the conservation of mass: 
\begin{gather}
\frac{\partial u}{\partial t} + (u \cdot \nabla) u = - \frac{\nabla P}{\rho} + \nu \nabla^2 u + g,
\label{eq: navier stokes momentum}
\\
\nabla \cdot u = 0.
\label{eq: navier stokes mass}
\end{gather}
In this context, $u$ is the velocity, $P$ is the pressure the fluid exerts, $\rho$ is the density of the fluid (usually assumed to be constant), $\nu$ is the kinematic viscosity coefficient that indicates the thickness of the fluid, and $g$ denotes the acceleration due to gravity. With this PDE, three data sets were created using a smoke and a liquid solver. For all data, 2D simulations were run until a certain step, and useful data fields were exported afterwards.

\subsubsection*{Smoke}
For the smoke data, a standard Eulerian fluid solver using a preconditioned pressure solver based on the conjugate gradient method and Semi-Lagrangian advection scheme was employed. 

\begin{figure*}[htp]
    \centering
    \subfigure[Data samples generated without noise: tiny output changes following generation ordering]{
        \includegraphics[width=0.99\textwidth]{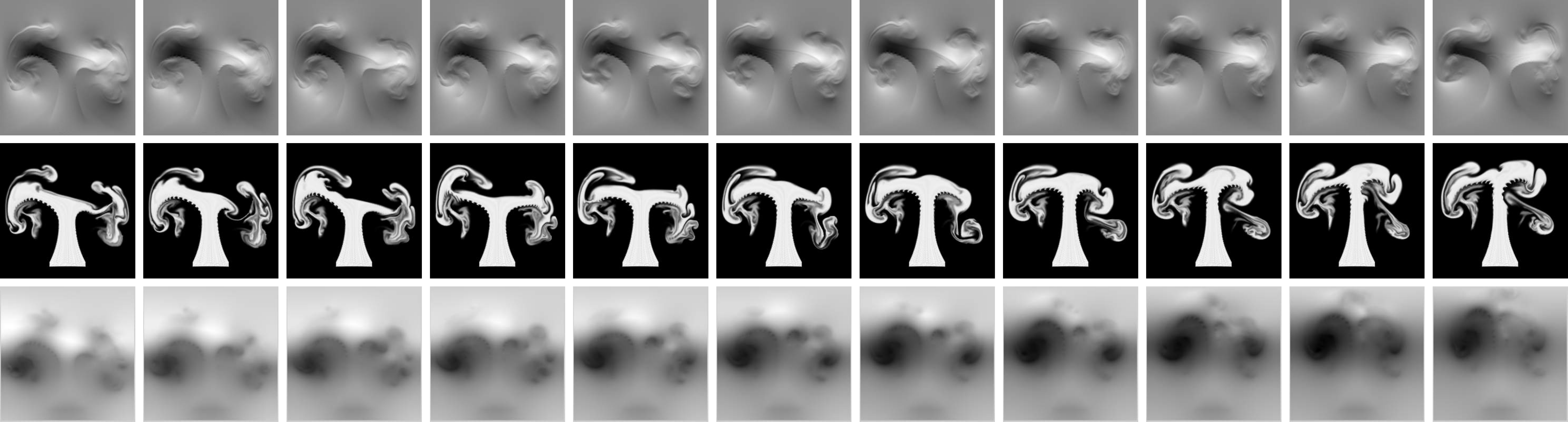} }
    ~\par\smallskip
    \subfigure[Normal training data samples with noise: larger output changes but ordering still applies]{
        \includegraphics[width=0.99\textwidth]{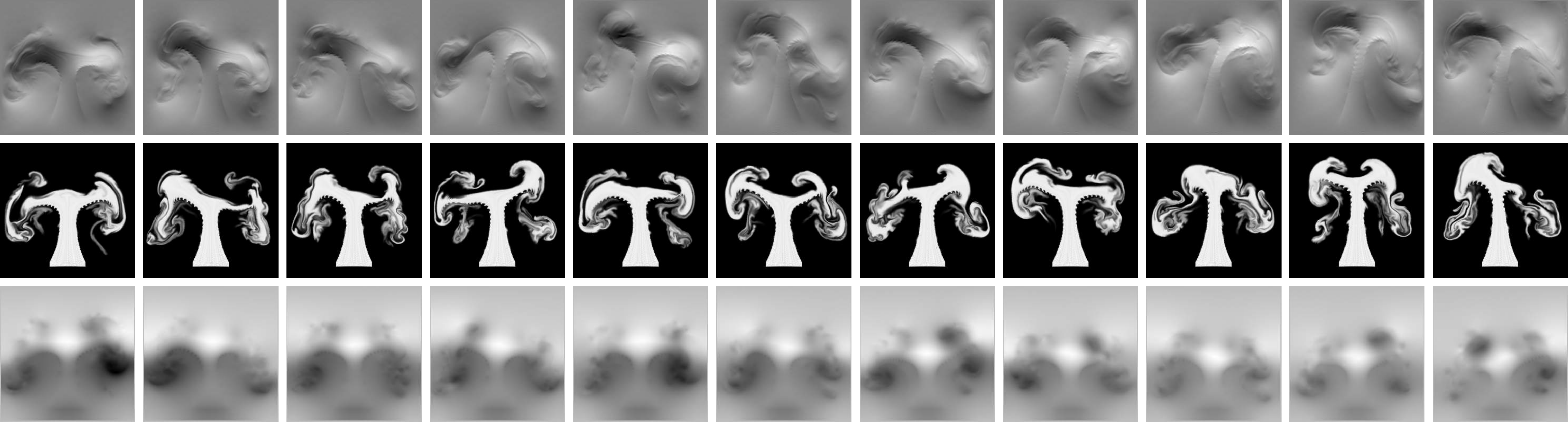} }
    ~\par\smallskip
    \subfigure[Outlier data samples: noise can override the generation ordering by chance]{
        \includegraphics[width=0.99\textwidth]{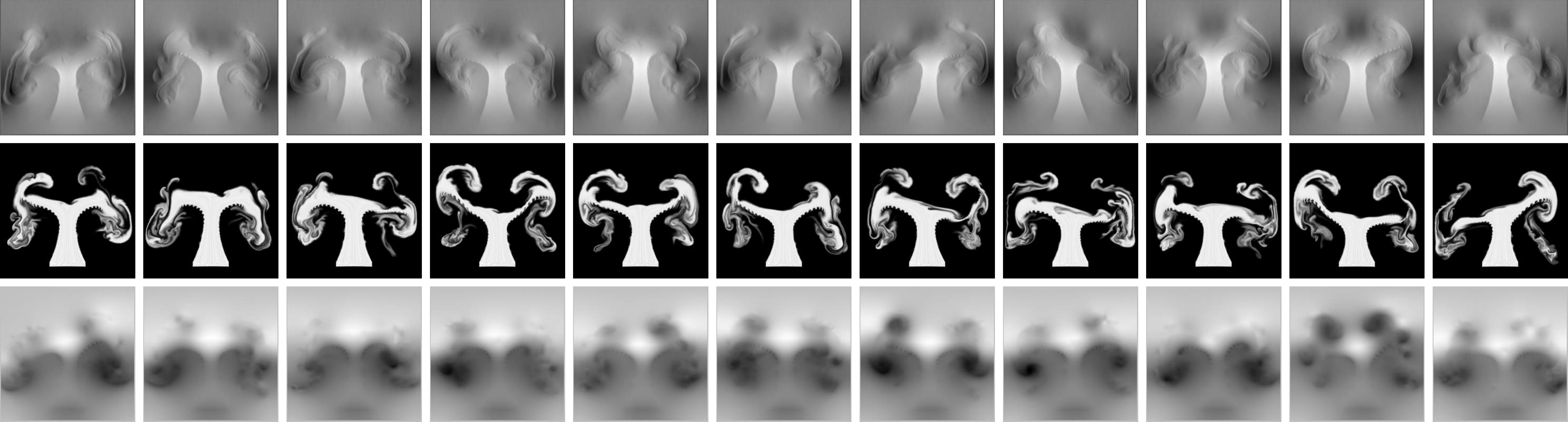} }
        
    \caption{Various smoke simulation examples using one component of the velocity (top rows), the density (middle rows), and the pressure field (bottom rows).}
    \label{fig: smoke example}
    \vfill
\end{figure*}

\begin{figure*}[htp]
    \centering
    \subfigure[Data samples generated without noise: tiny output changes following generation ordering]{
        \includegraphics[width=0.99\textwidth]{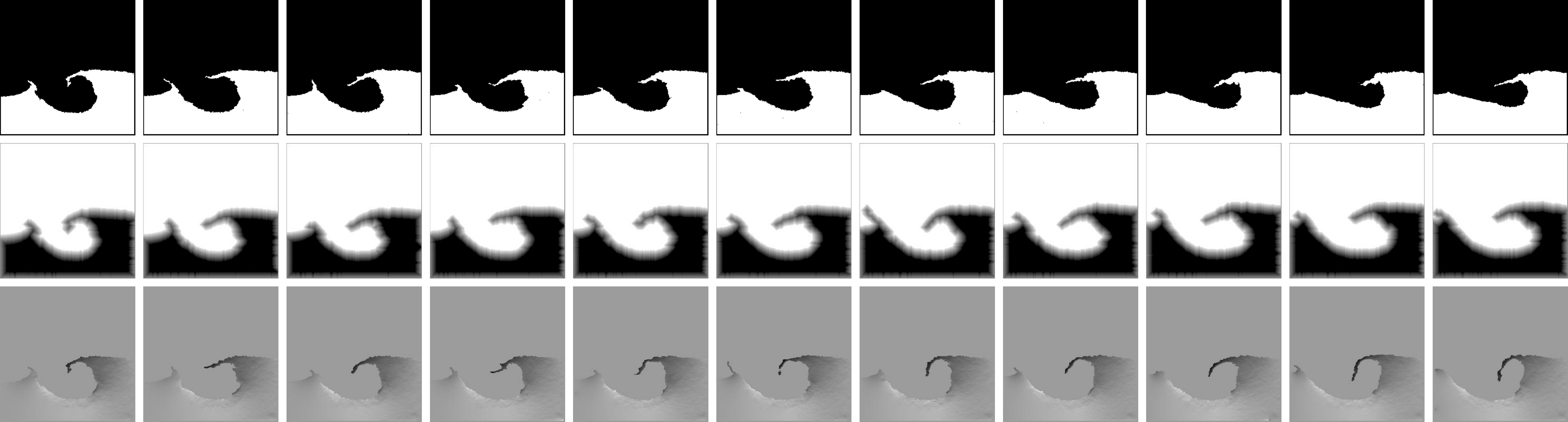} }
    ~\par\smallskip
    \subfigure[Normal training data samples with noise: larger output changes but ordering still applies]{
        \includegraphics[width=0.99\textwidth]{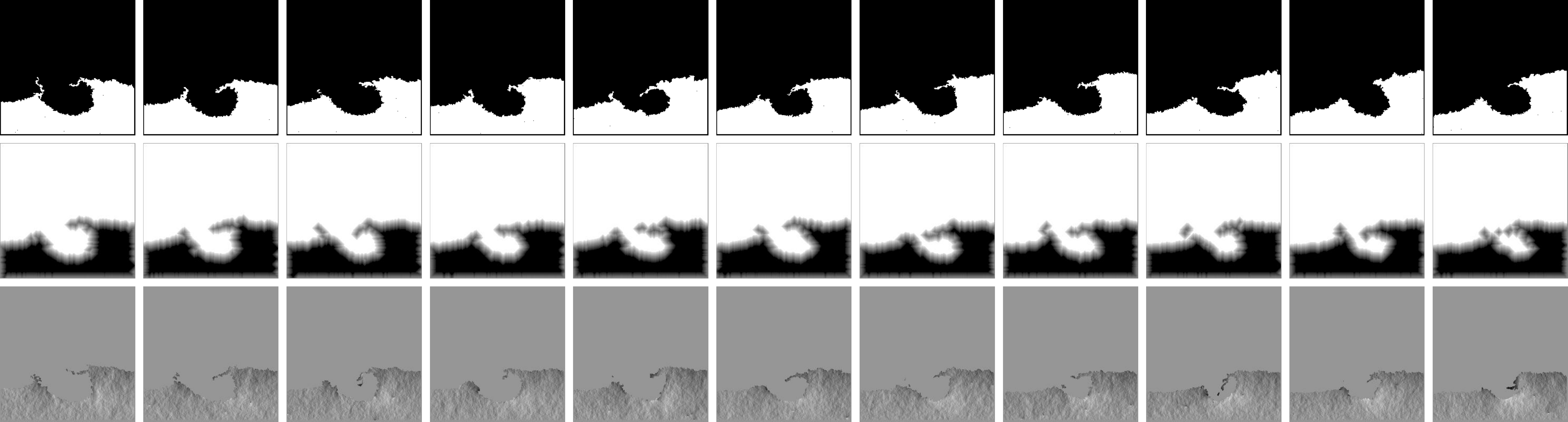} }
    ~\par\smallskip
    \subfigure[Outlier data samples: noise can override the generation ordering by chance]{
        \includegraphics[width=0.99\textwidth]{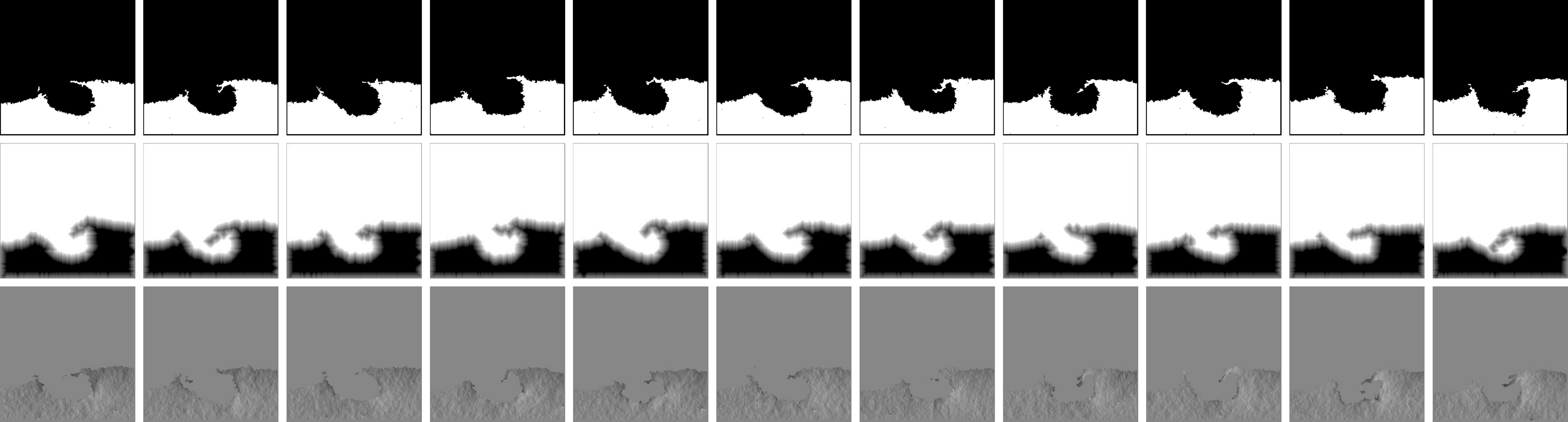} }
    ~\par\smallskip
    \subfigure[Data samples from test set with additional background noise]{
        \includegraphics[width=0.99\textwidth]{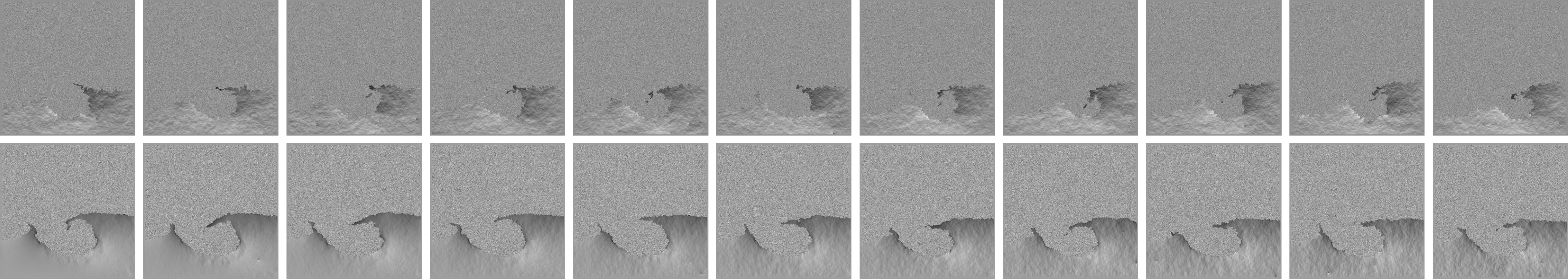}
        \label{fig: liquid example (test)} }
        
    \caption{Several liquid simulation examples using the binary indicator flags (top rows), the extrapolated level set values (middle rows), and one component of the velocity field (bottom rows) for the training data and only the velocity field for the test data.}
    \label{fig: liquid example}
    \vfill
\end{figure*}

\begin{figure*}[htp]
    \centering
    \subfigure[Data samples generated without noise: tiny output changes following generation ordering]{
        \includegraphics[width=0.99\textwidth]{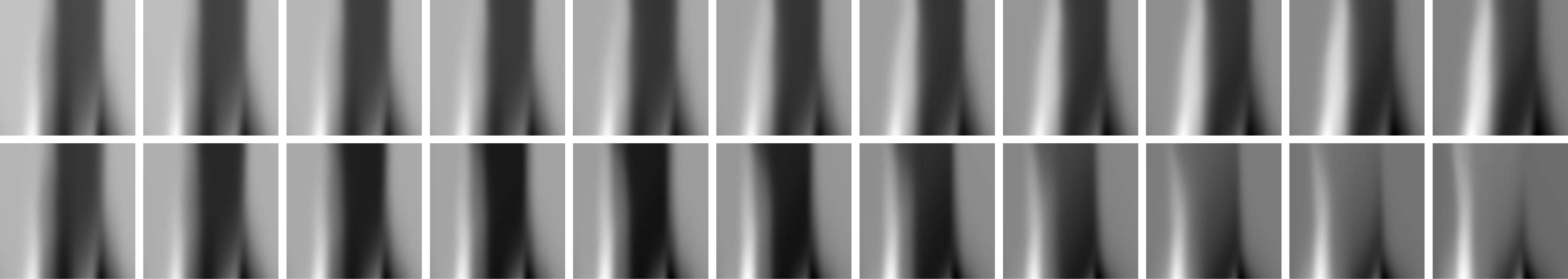} }
    ~\par\smallskip
    \subfigure[Normal training data samples with noise: larger output changes but ordering still applies]{
        \includegraphics[width=0.99\textwidth]{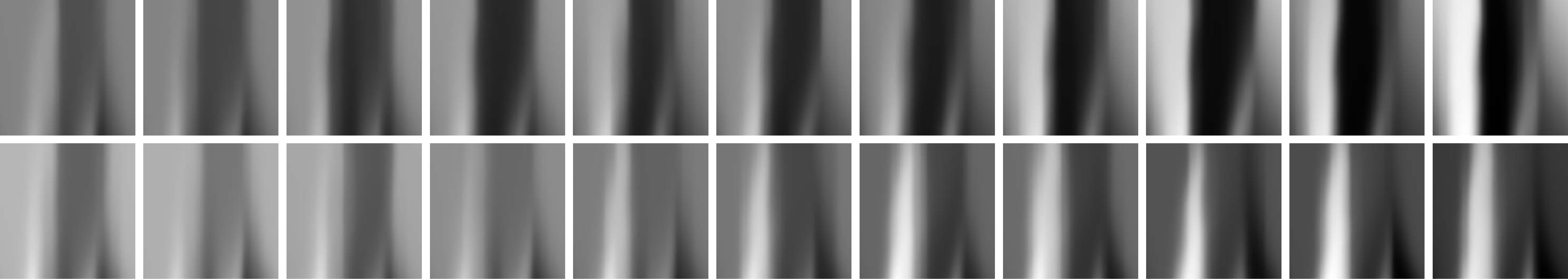} }
    ~\par\smallskip
    \subfigure[Outlier data samples: noise can override the generation ordering by chance]{
        \includegraphics[width=0.99\textwidth]{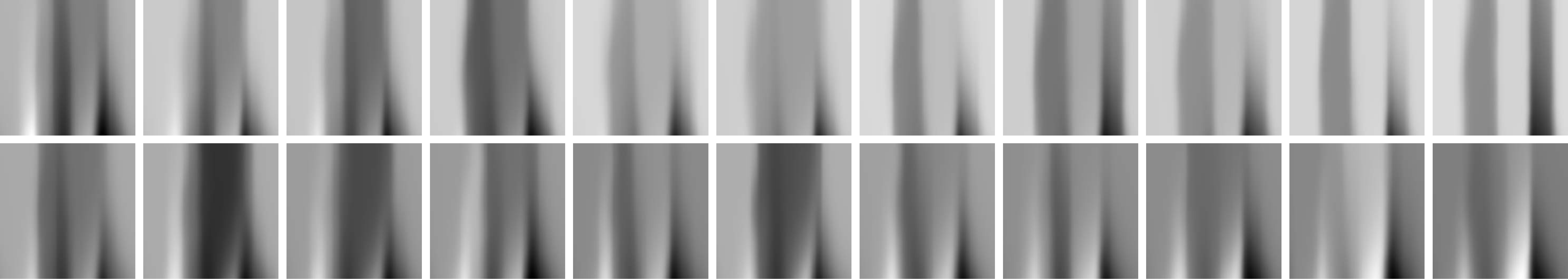} }
    ~\par\smallskip
    \subfigure[Data samples from test set with additional background noise]{
        \includegraphics[width=0.99\textwidth]{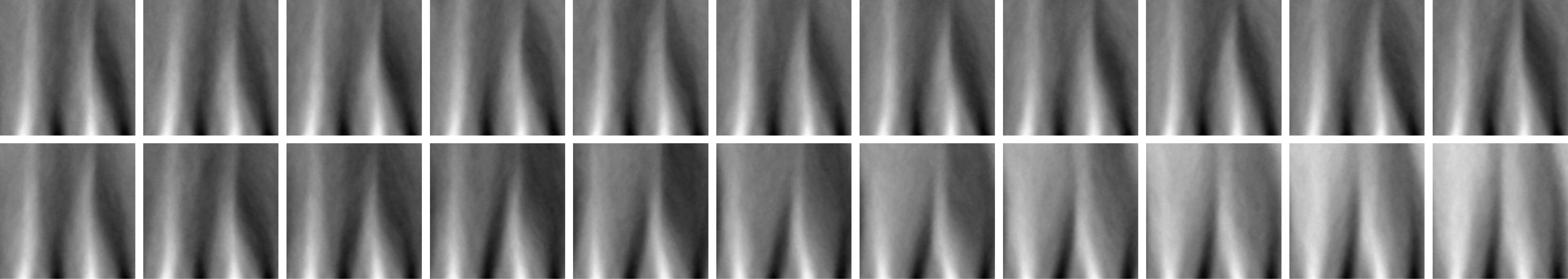}
        \label{fig: advDiff example (test)} }
        
    \caption{Various examples from the Advection-Diffusion equation using the density field.}
    \label{fig: advDiff example}
    \vfill
\end{figure*}

The general setup for every smoke simulation consists of a rectangular smoke source at the bottom with a fixed additive noise pattern to provide smoke plumes with more details. Additionally, there is a downwards directed, spherical force field area above the source, which divides the smoke in two major streams along it. We chose this solution over an actual obstacle in the simulation in order to avoid overfitting to a clearly defined black obstacle area inside the smoke data. Once the simulation reaches a predefined time step, the density, pressure, and velocity fields (separated by dimension) are exported and stored. Some example sequences can be found in Fig.~\ref{fig: smoke example}. With this setup, the following initial conditions were varied in isolation:
\vspace{-0.2cm}
\begin{itemize}
    \setlength\itemsep{0em}
    \item Smoke buoyancy in x- and y-direction
    \item Strength of noise added to the velocity field
    \item Amount of force in x- and y-direction provided by the force field
    \item Orientation and size of the force field
    \item Position of the force field in x- and y-direction
    \item Position of the smoke source in x- and y-direction
\end{itemize}

Overall, 768 individual smoke sequences were used for training, and the validation set contains 192 sequences with different initialization seeds.

\subsubsection*{Liquid}
For the liquid data, a solver based on the fluid implicit particle (FLIP) method \cite{zhu2005} was employed. It is a hybrid Eulerian-Lagrangian approach that replaces the Semi-Lagrangian advection scheme with particle based advection to reduce numerical dissipation. Still, this method is not optimal as we experienced problems such as mass loss, especially for larger noise values.

The simulation setup consists of a large breaking dam and several smaller liquid areas for more detailed splashes. After the dam hits the simulation boundary, a large, single drop of liquid is created in the middle of the domain that hits the already moving liquid surface. Then, the extrapolated level set values, binary indicator flags, and the velocity fields (separated by dimension) are saved.
Some examples are shown in Fig.~\ref{fig: liquid example}. The list of varied parameters include:
\vspace{-0.2cm}
\begin{itemize}
    \setlength\itemsep{0em}
    \item Radius of the liquid drop
    \item Position of the drop in x- and y-direction
    \item Amount of additional gravity force in x- and y-direction
    \item Strength of noise added to the velocity field
\end{itemize}

The liquid training set consists of 792 sequences and the validation set of 198 sequences with different random seeds. For the liquid test set, additional background noise was added to the velocity field of the simulations as displayed in Fig.~\ref{fig: liquid example (test)}. Because this only alters the velocity field, the extrapolated level set values and binary indicator flags are not used for this data set, leading to 132 sequences.

\subsection{Advection-Diffusion and Burger's Equation}
For these PDEs, our solvers only discretize and solve the corresponding equation in 1D. Afterwards, the different time steps of the solution process are concatenated along a new dimension to form 2D data with one spatial and one time dimension.

\begin{figure*}[ht]
    \centering
    \subfigure[Data samples generated without noise: tiny output changes following generation ordering]{
        \includegraphics[width=0.99\textwidth]{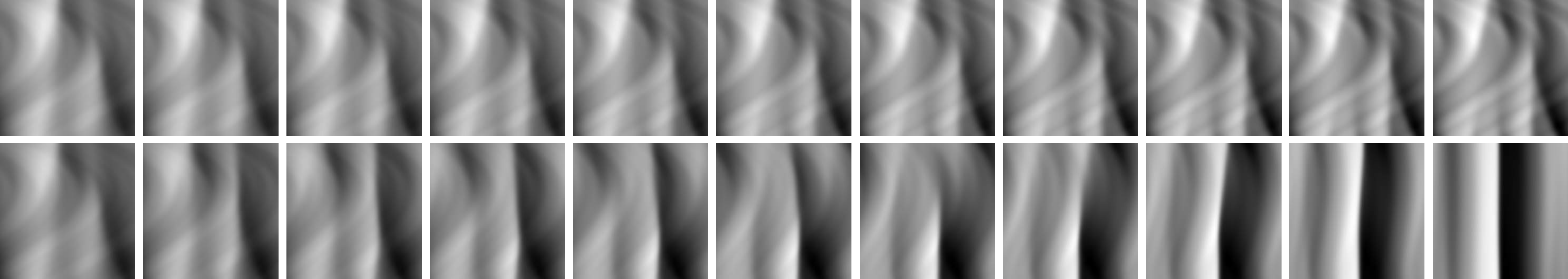} }
    ~\par\smallskip
    \subfigure[Normal training data samples with noise: larger output changes but ordering still applies]{
        \includegraphics[width=0.99\textwidth]{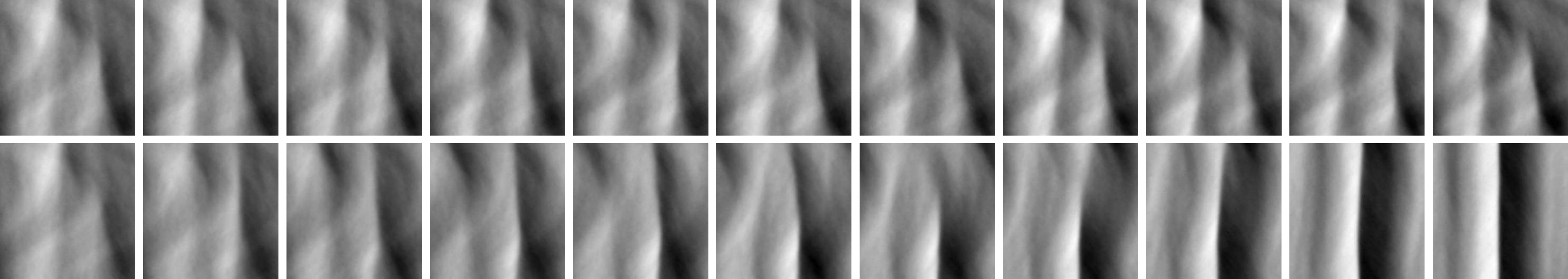} }
    ~\par\smallskip
    \subfigure[Outlier data samples: noise can override the generation ordering by chance]{
        \includegraphics[width=0.99\textwidth]{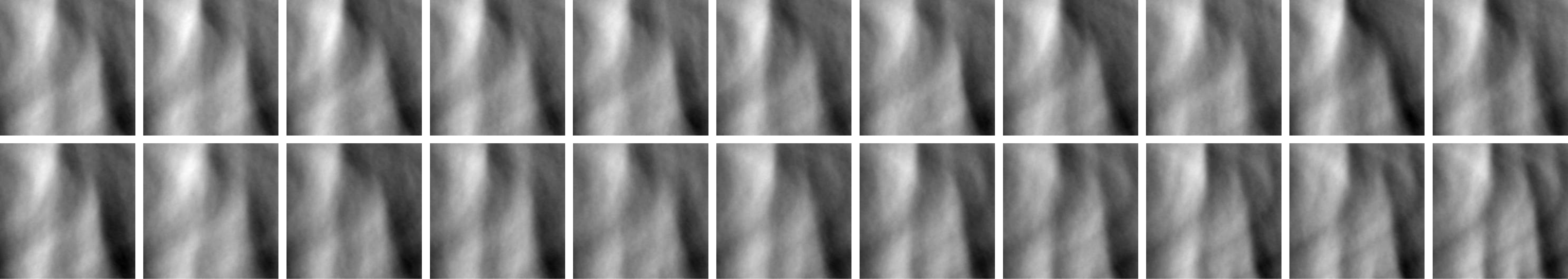} }
        
    \caption{Different simulation examples from the Burger's equation using the velocity field.}
    \label{fig: burgersEq example}
\end{figure*}

\subsubsection*{Advection-Diffusion Equation} This equation describes how a passive quantity is transported inside a velocity field due to the processes of advection and diffusion.
Eq.~(\ref{eq: advection diffusion}) is the simplified Advection-Diffusion equation with constant diffusivity and no sources or sinks.
\begin{equation}
\frac{\partial d}{\partial t} = \nu \nabla^2 d - u \cdot \nabla d,
\label{eq: advection diffusion}
\end{equation}
where $d$ denotes the density, $u$ is the velocity, and $\nu$ is the kinematic viscosity (also known as diffusion coefficient) that determines the strength of the diffusion. 
Our solver employed a simple implicit time integration and a diffusion solver based on conjugate gradient without preconditioning. The initialization for the 1D fields of the simulations was created by overlaying multiple parameterized sine curves with random frequencies and magnitudes. 

In addition, continuous forcing controlled by further parameterized sine curves was included in the simulations over time. In this case, the only initial conditions to vary are the forcing and initialization parameters of the sine curves and the strength of the added noise. From this PDE, only the passive density field was used as shown in Fig.~\ref{fig: advDiff example}. Overall, 798 sequences are included in the training set and 190 sequences with a different random initialization in the validation set.

For the Advection-Diffusion test set, the noise was instead added directly to the passive density field of the simulations. This results in 190 sequences with more small scale details as shown in Fig.~\ref{fig: advDiff example (test)}.

\subsubsection*{Burger's Equation}
This equation is very similar to the Advection-Diffusion equation and describes how the velocity field itself changes due to diffusion and advection:
\begin{equation}
\frac{\partial u}{\partial t} = \nu \nabla^2 u - u \cdot \nabla u.
\label{eq: burgers}
\end{equation}
Eq.~(\ref{eq: burgers}) is known as the viscous form of the Burger's equation that can develop shock waves, and again $u$ is the velocity and $\nu$ denotes the kinematic viscosity. Our solver for this PDE used a slightly different implicit time integration scheme, but the same diffusion solver as used for the Advection-Diffusion equation.

The simulation setup and parameters were also the same; the only difference is that the velocity field instead of the density is exported. As a consequence, the data in Fig.~\ref{fig: burgersEq example} looks relatively similar to those from the Advection-Diffusion equation. The training set features 782 sequences, and the validation set contains 204 sequences with different random seeds.

\subsection{Other Data-Sets}
The remaining data sets are not based on PDEs and thus not generated with the proposed method. The data is only used to test the generalization of the discussed metrics and not for training or validation. The Shapes test set contains 160 sequences, the Video test set consists 131 sequences, and the TID test set features 216 sequences.

\begin{figure*}[bp]
    \centering
    \includegraphics[width=0.99\textwidth]{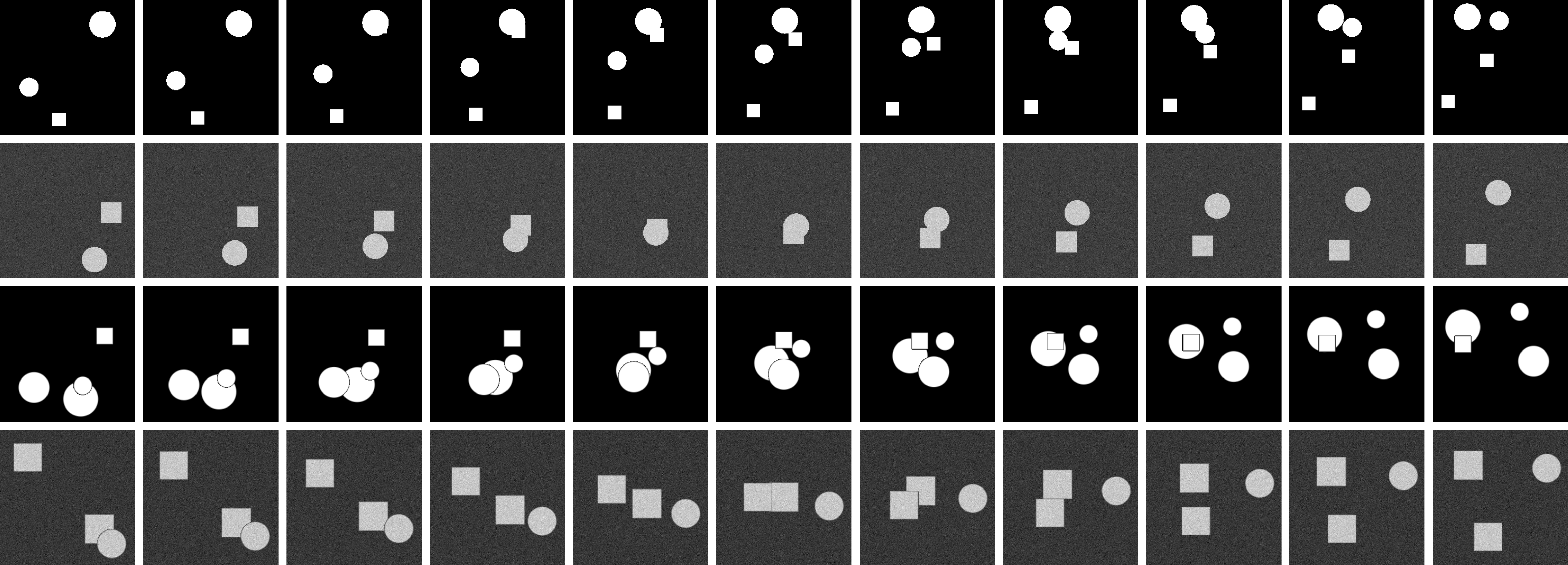}
    \caption{Examples from the shapes data set using a field with only binary shape values (first row), shape values with additional noise (second row), smoothed shape values (third row), and smoothed values with additional noise (fourth row).}
    \label{fig: shapes example}
    \vfill
\end{figure*}

\subsubsection*{Shapes}
This data set tests if the metrics are able to track simple, moving geometric shapes. To create it, a straight path between two random points inside the domain is generated and a random shape is moved along this path in steps of equal distance. The size of the used shape depends on the distance between the start and end point such that a significant fraction of the shape overlaps between two consecutive steps. It is also ensured that no part of the shape leaves the domain at any step by using a sufficiently big boundary area when generating the path.

With this method, multiple random shapes for a single data sample are produced, and their paths can overlap such that they occlude each other to provide an additional challenge. All shapes are moved in their parametric representation, and only when exporting the data, they are discretized onto a fixed binary grid. To add more variations to this simple approach, we also apply them in a non-binary way with smoothed edges and include additive Gaussian noise over the entire domain. Examples are shown in Fig.~\ref{fig: shapes example}.

\subsubsection*{Video}
\begin{figure*}[ht]
    \centering
    \includegraphics[width=0.93\textwidth]{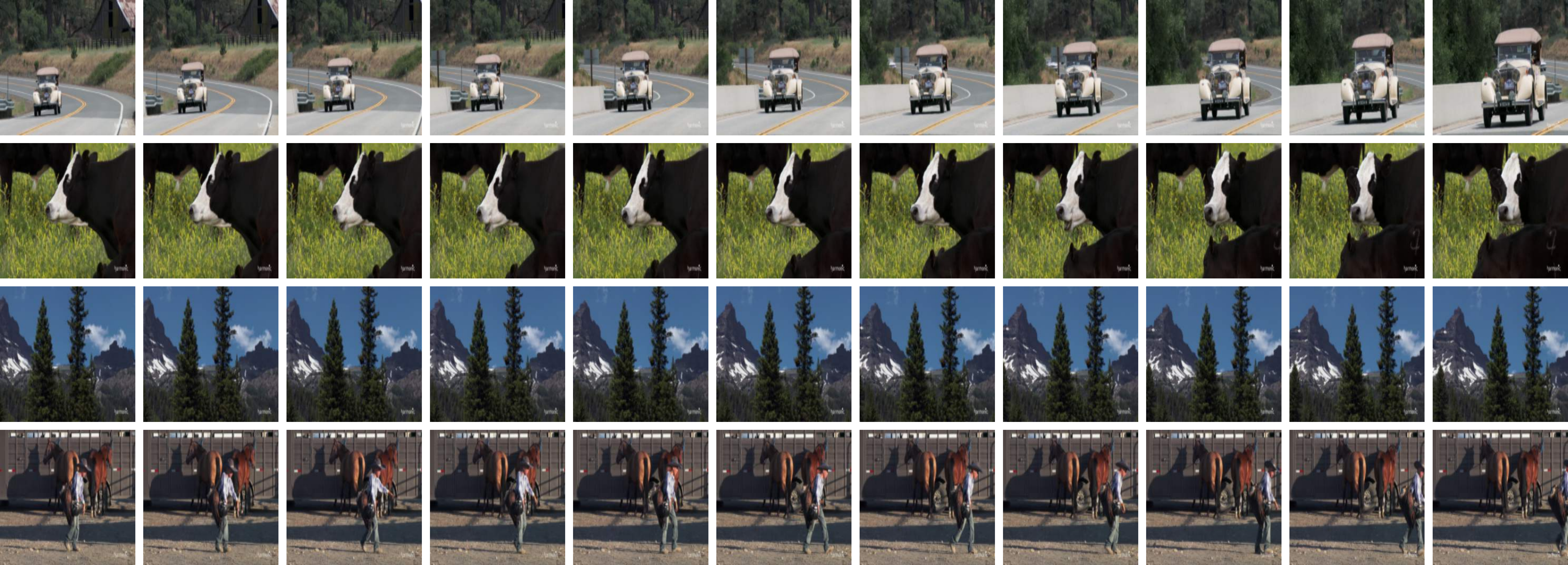}
    \caption{Multiple examples from the video data set.}
    \label{fig: video example}
\end{figure*}

For this data set, different publicly available video recordings were acquired and processed in three steps. 
First, videos with abrupt cuts, scene transitions, or camera movements were discarded, and afterwards the footage was broken down into single frames. Then, each frame was resized to match the spatial size of our other data by linear interpolation. Since directly using consecutive frames is no challenge for any analyzed metric and all of them recovered the ordering almost perfectly, we achieved a more meaningful data set by skipping several intermediate frames. For the final data set, we defined the first frame of every video as the reference and collected subsequent frames in an interval step of ten frames as the increasingly different variations. Some data examples can be found in Fig.~\ref{fig: video example}.

\begin{figure*}[ht]
    \centering
    \includegraphics[width=0.93\textwidth]{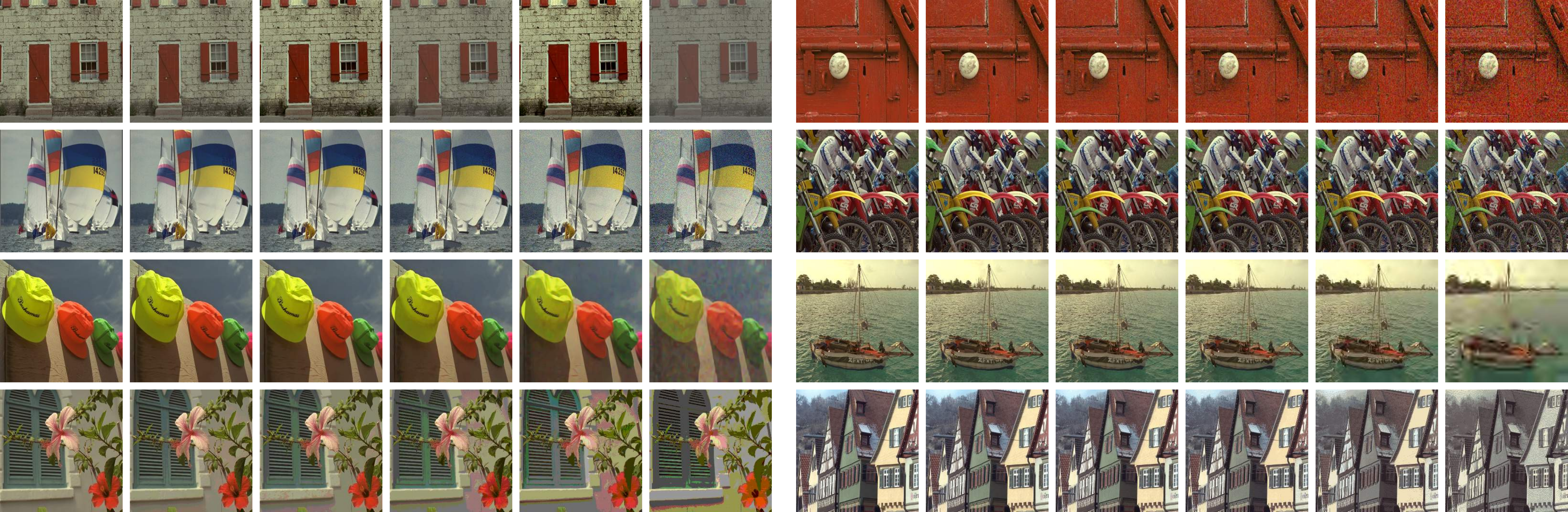}
    \caption{Examples from the TID2013 data set proposed by \citeauthor{ponomarenko2015}. Displayed are a change of contrast, three types of noise, denoising, jpg2000 compression, and two color quantizations (from left to right and top to bottom).}
    \label{fig: TID example}
\end{figure*}

\subsubsection*{TID2013}
This data set was created by \citeauthor{ponomarenko2015} and used without any further modifications. It consists of 25 reference images with 24 distortion types in five levels. As a result, it is not directly comparable to our data sets; thus, it is excluded from the test set aggregations. The distortions focus on various types of noise, image compression, and color changes. Fig.~\ref{fig: TID example} contains examples from the data set.

\subsection{Hardware}
Data generation, training, and metric evaluations were performed on a machine with an Intel i7-6850 (3.60Ghz) CPU and an NVIDIA GeForce GTX 1080 Ti GPU.

\section{Real-World Data}  \label{append: real-world data}
Below, we give details of the three data sets used for the evaluation in Section~\ref{subsec: real-world evaluation} of the main paper.

\subsection{ScalarFlow}
\begin{figure*}[hp]
    \centering
    \includegraphics[width=0.99\textwidth]{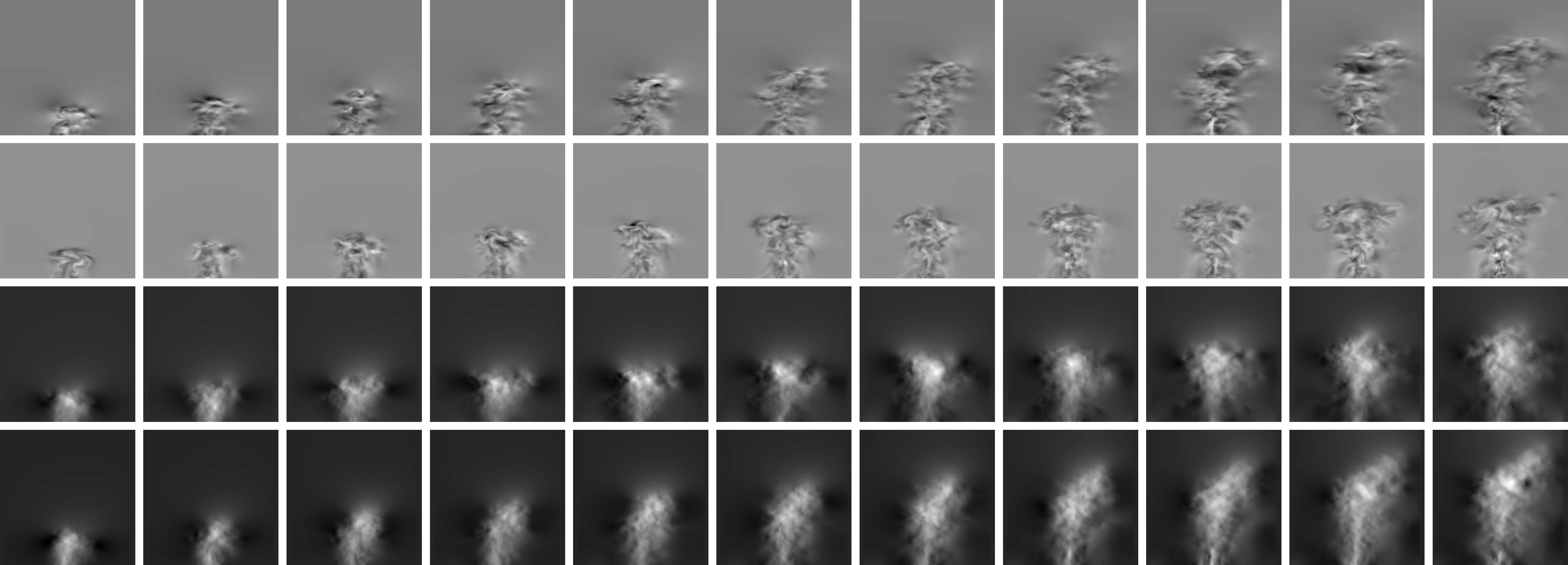}
    \caption{Four different smoke plume examples of the processed \textit{ScalarFlow} data set using one of the three velocity components. The two top rows show the center slice, and the two bottom rows show the mean along the z-dimension. The temporal interval between each image is ten simulation time steps.}
    \label{fig: scalarFlow example}
\end{figure*}

The \textit{ScalarFlow} data set \cite{eckert2019} contains 3D velocities of real-world scalar transport flows reconstructed from multiple camera perspectives. For our evaluation, we cropped the volumetric $100\times178\times100$ grids to $100\times160\times100$ such that they only contain the area of interest and convert them to 2D with two variants: either by using the center slice or by computing the mean along the z-dimension. Afterwards, the velocity vectors are split by channels, linearly interpolated to $256\times256$, and then normalized. Variations for each reconstructed plume are acquired by using frames in equal temporal intervals. We employed the velocity field reconstructions from 30 plumes (with simulation IDs $0-29$) for both compression methods. Fig.~\ref{fig: scalarFlow example} shows some example sequences.

\begin{figure*}[hp]
    \centering
    \includegraphics[width=0.99\textwidth]{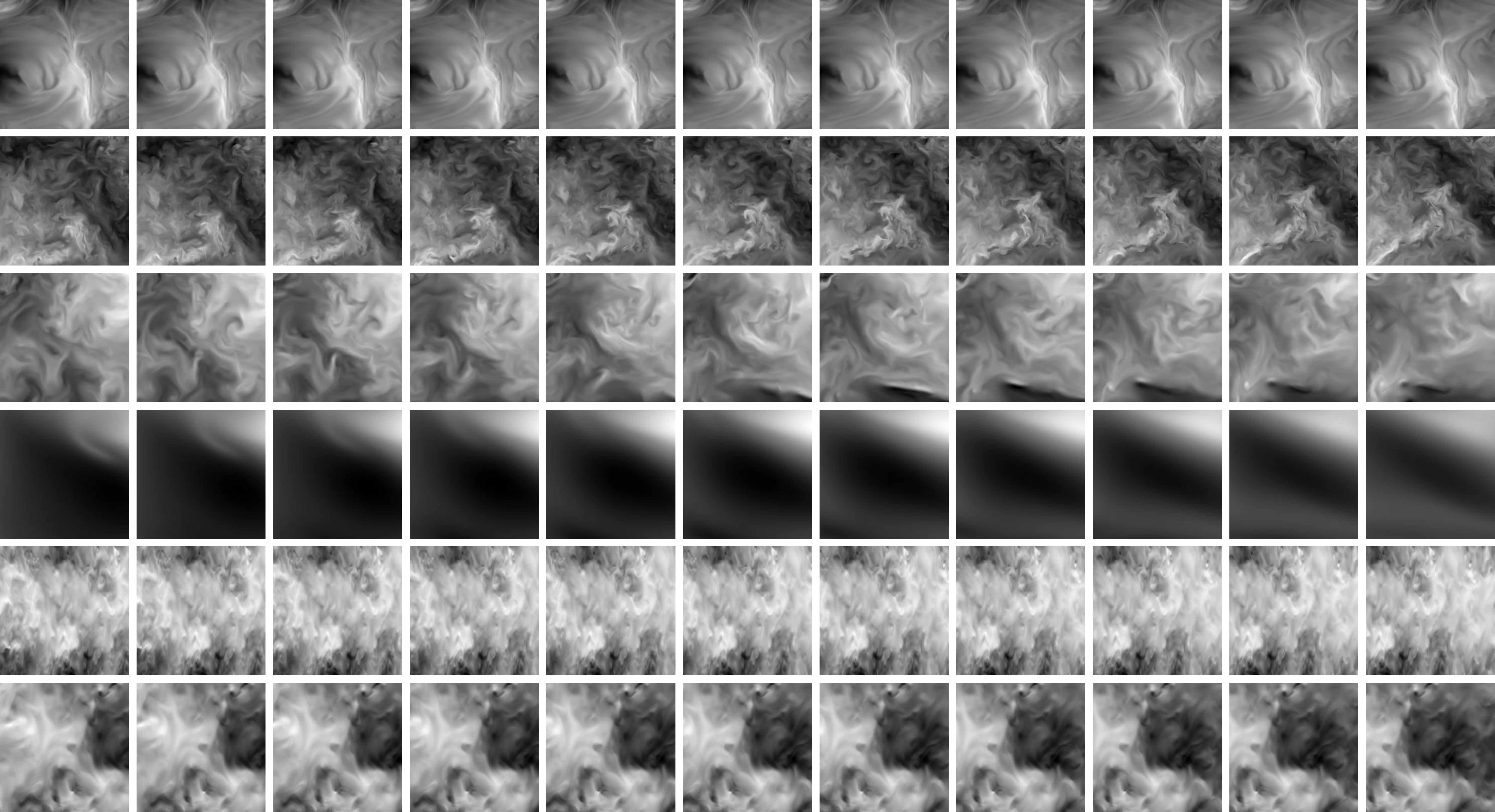}
    \caption{Data samples extracted from the Johns Hopkins Turbulence Database with a spatial or temporal interval of ten using one of the three velocity components. From top to bottom: \textit{mhd1024} and \textit{isotropic1024coarse} (varied time step), \textit{isotropic4096} and \textit{rotstrat4096} (varied z-position), \textit{channel} and \textit{channel5200} (varied x-position).}
    \label{fig: jhtdb example}
\end{figure*}

\subsection{Johns Hopkins Turbulence Database}
The Johns Hopkins Turbulence Database (\textit{JHTDB}) \citep{perlman2007} features various data sets of 3D turbulent flow fields created with direct numerical simulations (DNS). Here, we used three forced isotropic turbulence data sets with different resolutions (\textit{isotropic1024coarse}, \textit{isotropic1024fine}, and \textit{isotropic4096}), two channel flows with different Reynolds numbers (\textit{channel} and \textit{channel\-5200}), the forced magneto-hydrodynamic isotropic turbulence data set (\textit{mhd1024}), and the rotating stratified turbulence data set (\textit{rotstrat4096}).

For the evaluation, five $256\times256$ reference slices in the x/y-plane from each of the seven data sets are used. The spatial and temporal position of each slice is randomized within the bounds of the corresponding simulation domain. We normalize the value range and split the velocity vectors by component for an individual evaluation. Variants for each reference are created by gradually varying the x- and z-position of the slice in equal intervals. The temporal position of each slice is varied as well if a sufficient amount of temporally resolved data is available (for \textit{isotropic1024coarse}, \textit{isotropic1024fine}, \textit{channel}, and \textit{mhd1024}). This leads to 216 sequences in total. Fig.~\ref{fig: jhtdb example} shows examples from six of the \textit{JHTDB} data sets.

\begin{figure*}[htp]
    \centering
    \includegraphics[width=0.98\textwidth]{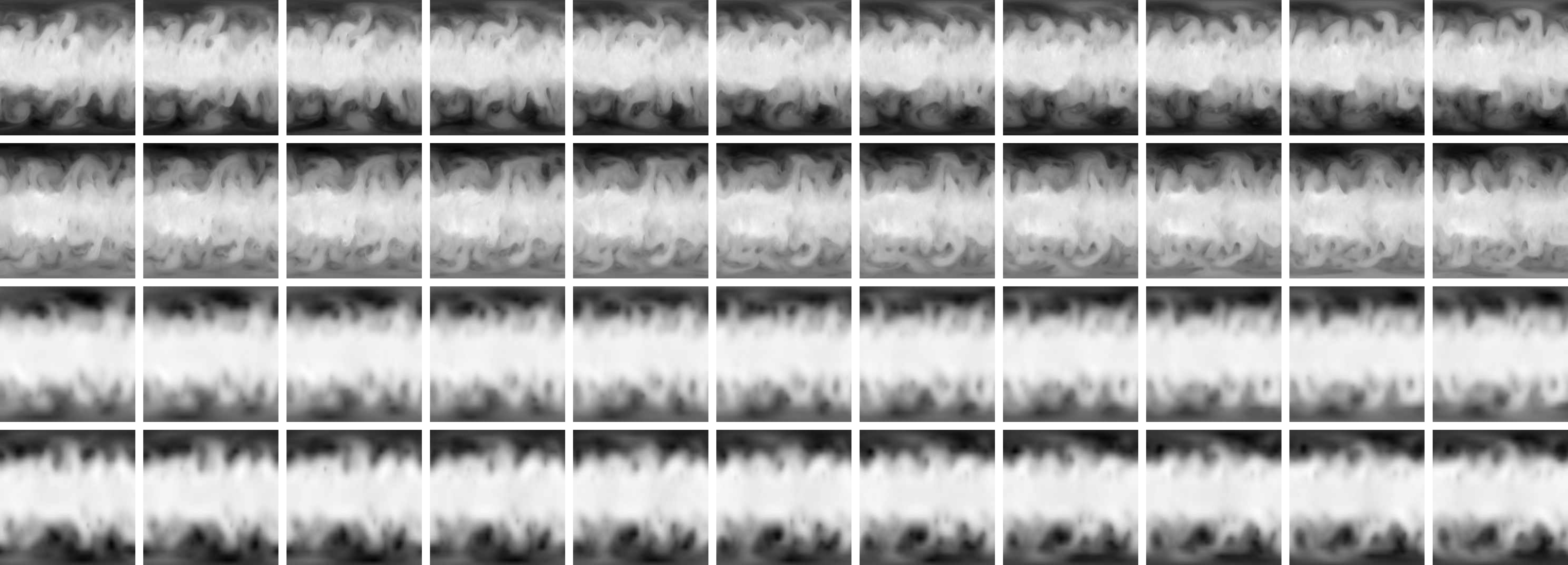}
    \caption{Examples of the processed \textit{WeatherBench} data: high-res temperature data \textit{1.40625deg/temperature} (upper two rows) and low-res geopotential data \textit{5.625deg/geopotential\_500} (lower two rows). The temporal interval spacing between the images is twenty hours.}
    \label{fig: weatherBench example}
\end{figure*}

\subsection{WeatherBench}
The \textit{WeatherBench} repository \cite{rasp2020} represents a collection of various weather measurements of different atmospherical quantities such as precipitation, cloud coverage, wind velocities, geopotential, and temperature. The data ranges from 1979 to 2018 with a fine temporal resolution and is stored on a Cartesian latitude-longitude grid of the earth. In certain subsets of the data, an additional dimension such as altitude or pressure levels is available. As all measurements are available as scalar fields, only a linear interpolation to the correct input size and a normalization was necessary in order to prepare the data.
We used the low-resolution geopotential data set at 500hPa (i.e., at around 5.5km height) with a size of $32\times64$ yielding smoothly changing features when upsampling the data. 
In addition, the high-res temperature data with a size of $128\times256$ for small scale details was used. For the temperature field, we used the middle atmospheric pressure level at 850hPa corresponding to an altitude of 1.5km in our experiments.

\begin{figure*}[ht]
    \centering
    \includegraphics[width=0.99\textwidth]{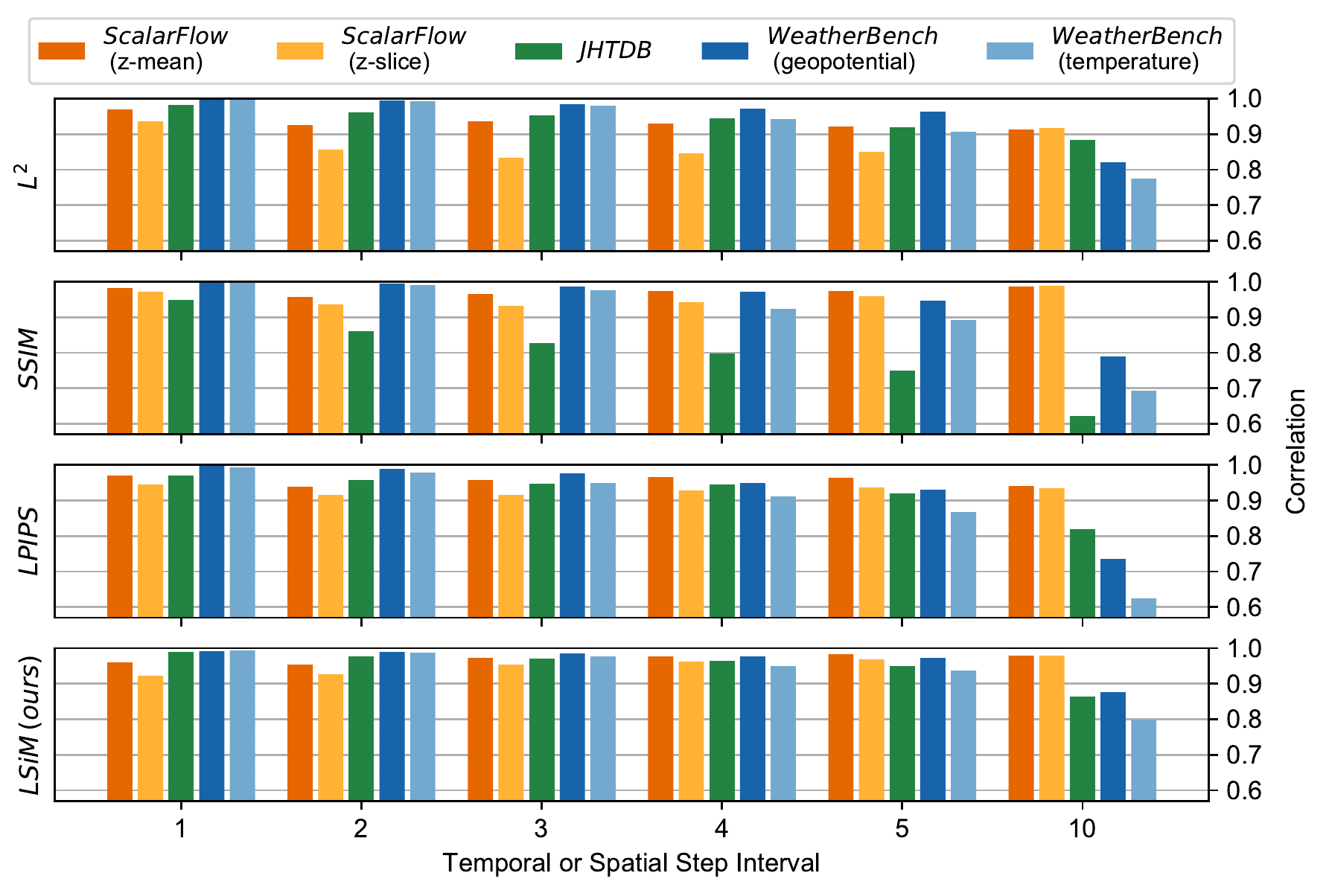}
    \vspace{-0.5cm}
    \caption{Detailed breakdown of the results when evaluating \textit{LSiM} on the individual data sets of \textit{ScalarFlow} (30 sequences each), \textit{JHTDB} (90 sequences each), and \textit{WeatherBench} (120 sequences each) with different step intervals.}
    \label{fig: real world evaluation detailed}
\end{figure*}

To create sequences with variations for a single time step of the weather data, we used frames in equal time intervals, similar to the \textit{ScalarFlow} data. Due to the very fine temporal discretization of the data, we only use a temporal interval of two hours as the smallest interval step of one in Fig.~\ref{fig: real world evaluation detailed}. We sampled three random starting points in time from each of the 40 years of measurements, resulting in 120 temperature and geopotential sequences overall. Fig.~\ref{fig: weatherBench example} shows a collection of example sequences.

\subsection{Detailed Results}
For each of the variants explained in the previous sections, we create test sets with six different spatial and temporal intervals. 
Fig.~\ref{fig: real world evaluation detailed} shows the combined Spearman correlation of the sequences for different interval spacings when evaluating various metrics. For the results in Fig.~\ref{fig: real world evaluation} in the main paper, all correlation values shown here are aggregated by data source via mean and standard deviation.

While our metric reliably recovers the increasing distances within the data sets, the individual measurements exhibit interesting differences in terms of their behavior for varying distances. As \textit{JHTDB} and \textit{WeatherBench} contain relatively uniform phenomena, a larger step interval creates more difficult data as the simulated and measured states contain changes that are more and more difficult to analyze along a sequence.
For \textit{ScalarFlow}, on the other hand, the difficulty decreases for larger intervals due to the large-scale motion of the reconstructed plumes. As a result of buoyancy forces, the observed smoke rises upwards into areas where no smoke has been before. For the network, this makes predictions relatively easy as the large-scale translations are indicative of the temporal progression, and small scale turbulence effects can be largely ignored. For this data set, smaller intervals are more difficult as the overall shape of the plume barely changes while the complex evolution of small scale features becomes more important.

Overall, the \textit{LSiM} metric recovers the ground truth ordering of the sequences very well as indicated by the consistently high correlation values in Fig.~\ref{fig: real world evaluation detailed}. The other metrics comes close to these results on certain sub-datasets but are significantly less consistent. \textit{SSIM} struggles on \textit{JHTDB} across all interval sizes, and \textit{LPIPS} cannot keep up on \textit{WeatherBench}, especially for larger intervals. $L^2$ is more stable overall, but consistently stays below the correlation achieved by \textit{LSiM}.


\section{Additional Evaluations} \label{append: additional evaluations}
In the following, we demonstrate other ways to compare the performance of the analyzed metrics on our data sets. In Tab.~\ref{table: results pearson}, the Pearson correlation coefficient is used instead of Spearman's rank correlation coefficient. While Spearman's correlation measures monotonic relationships by using ranking variables, it directly measures linear relationships.

The results in Tab.~\ref{table: results pearson} match very closely to the values computed with Spearman's rank correlation coefficient. The best performing metrics in both tables are identical; only the numbers slightly vary. Since a linear and a monotonic relation describes the results of the metrics similarly well, there are no apparent non-linear dependencies that cannot be captured using the Pearson correlation.

\begin{table*}[p]
    \caption{Performance comparison on validation and test data sets measured in terms of the Pearson correlation coefficient of ground truth against predicted distances. \best{Bold+underlined} values show the best performing metric for each data set, \bestErr{bold} values are within a $0.01$ error margin of the best performing, and \bad{italic} values are $0.2$ or more below the best performing. On the right a visualization of the combined test data results is shown for selected models.
    }
    \label{table: results pearson}
    \vspace{0.1cm}
    
    \centering
    \small
    \begin{tabular}[b]{l c c c c | c | c c c c c}
        \toprule
        \multirow{2}{*}[-1.3mm]{\bf Metric} & \multicolumn{4}{c |}{\bf Validation data sets} & \multicolumn{6}{c}{\bf Test data sets} \\
        \cmidrule(lr){2-5} \cmidrule(lr){6-11}
        & \texttt{Smo} & \texttt{Liq} & \texttt{Adv} & \texttt{Bur} & \texttt{TID} & \texttt{LiqN} & \texttt{AdvD} & \texttt{Sha} & \texttt{Vid} & \texttt{All} \\
        \cmidrule(lr){1-11}

        \it $L^2$                             & 0.66 & 0.80 & 0.72 & 0.60  & 0.82 & 0.73 & 0.55 & \bad{0.66} & 0.79 & 0.60 \\
        \it SSIM                              & 0.69 & 0.74 & 0.76 & 0.70  & 0.78 & \bad{0.26} & \best{0.69} & \bad{0.49} & 0.73 & 0.53 \\
        \it LPIPS v0.1.                       & 0.63 & 0.68 & 0.66 & 0.71  & \bestErr{0.85} & \bad{0.49} & 0.61 & 0.84 & \bestErr{0.83} & 0.65 \\
        \cmidrule(lr){2-11}

        \it AlexNet\textsubscript{random}     & 0.63 & 0.69 & 0.67 & 0.65  & 0.83 & 0.64 & 0.63 & 0.74 & 0.81 & 0.65 \\
        \it AlexNet\textsubscript{frozen}     & 0.66 & 0.69 & 0.68 & 0.71  & \bestErr{0.85} & \bad{0.39} & 0.61 & 0.86 & \bestErr{0.83} & 0.64 \\
        \it Optical flow                      & 0.63 & \bad{0.56} & \bad{0.37} & \bad{0.39}  & \bad{0.49} & \bad{0.45} & \bad{0.28} & \bad{0.61} & 0.74 & \bad{0.48} \\
        \it Non-Siamese                       & 0.77 & \best{0.84} & 0.78 & \bestErr{0.74}  & 0.67 & \best{0.81} & 0.64 & \bad{0.27} & 0.79 & 0.60 \\
        \it Skip\textsubscript{from scratch}  & \best{0.79} & \bestErr{0.83} & \best{0.80} & \bestErr{0.73}  & \bestErr{0.85} & 0.78 & 0.61 & 0.79 & \best{0.84} & \bestErr{0.71} \\
        \cmidrule(lr){2-11}
        
        \it LSiM\textsubscript{noiseless}     & 0.77 & 0.77 & 0.76 & 0.72  & \bestErr{0.86} & 0.62 & 0.58 & 0.84 & \bestErr{0.83} & 0.68 \\
        \it LSiM\textsubscript{strong noise}  & 0.65 & \bad{0.64} & 0.66 & 0.68  & 0.81 & \bad{0.39} & 0.53 & \best{0.90} & 0.82 & 0.64 \\
        \it LSiM (ours)                       & \bestErr{0.78} & 0.82 & \bestErr{0.79} & \best{0.74}  & \best{0.86} & 0.79 & 0.58 & 0.87 & 0.82 & \best{0.72} \\
   
        \bottomrule
    \end{tabular}
    \includegraphics[width=0.156\textwidth]{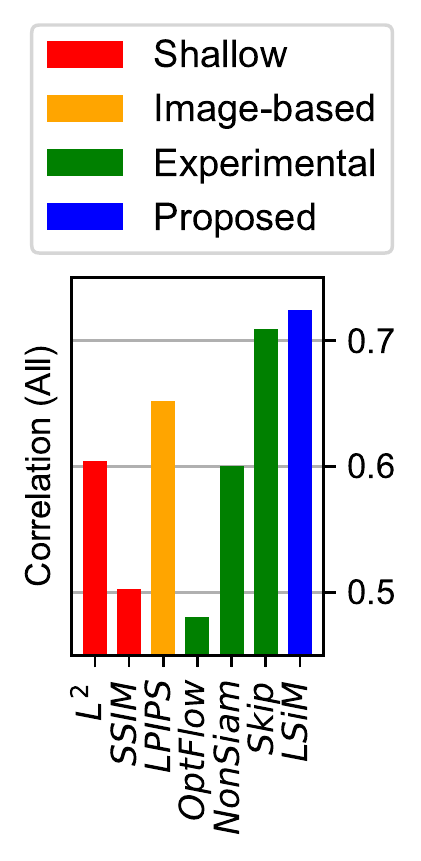}
\end{table*}

\begin{table*}[p]
    \vspace{-0.1cm}
    \caption{Performance comparison on validation data sets measured by computing mean and standard deviation (in brackets) of Pearson correlation coefficients (ground truth against predicted distances) from individual data samples. \best{Bold+underlined} values show the best performing metric for each data set, \bestErr{bold} values are within a $0.01$ error margin of the best performing, and \bad{italic} values are $0.2$ or more below the best performing. On the right a visualization of the combined test data results is shown for selected models.}
    \label{table: results pearson mean 1}
    \vspace{0.1cm}
    
    \centering
    \small
    \begin{tabular}[b]{l c c c c}
        \toprule
        \multirow{2}{*}[-1.3mm]{\bf Metric} & \multicolumn{4}{c}{\bf Validation data sets} \\
        \cmidrule(lr){2-5}
        & \texttt{Smo} & \texttt{Liq} & \texttt{Adv} & \texttt{Bur} \\
        \cmidrule(lr){1-5}

        \it $L^2$                             & 0.68 (0.27) & 0.82 (0.18) & 0.74 (0.24) & 0.63 (0.33) \\
        \it SSIM                              & 0.71 (0.23) & 0.75 (0.23) & 0.79 (0.21) & 0.73 (0.33) \\
        \it LPIPS v0.1.                       & 0.66 (0.29) & 0.71 (0.24) & 0.70 (0.29) & 0.75 (0.28) \\
        \cmidrule(lr){2-5}

        \it AlexNet\textsubscript{random}     & 0.65 (0.28) & 0.71 (0.29) & 0.71 (0.27) & 0.68 (0.31) \\
        \it AlexNet\textsubscript{frozen}     & 0.69 (0.27) & 0.72 (0.25) & 0.71 (0.27) & 0.74 (0.29) \\
        \it Optical flow                      & 0.66 (0.38) & \bad{0.59 (0.47)} & \bad{0.38 (0.52)} & \bad{0.41 (0.49)} \\
        \it Non-Siamese                       & 0.80 (0.19) & \best{0.87 (0.14)} & \bestErr{0.81 (0.20)} & 0.76 (0.32) \\
        \it Skip\textsubscript{from scratch}  & \best{0.81 (0.19)} & 0.85 (0.16) & \best{0.82 (0.19)} & \bestErr{0.77 (0.30)} \\
        \cmidrule(lr){2-5}
        
        \it LSiM\textsubscript{noiseless}     & 0.79 (0.21) & 0.79 (0.20) & 0.79 (0.23) & 0.76 (0.29) \\
        \it LSiM\textsubscript{strong noise}  & 0.67 (0.28) & \bad{0.66 (0.29)} & 0.68 (0.30) & 0.70 (0.32) \\
        \it LSiM (ours)                       & \bestErr{0.81 (0.20)} & 0.84 (0.16) & \bestErr{0.81 (0.19)} & \best{0.78 (0.28)} \\
   
        \bottomrule
    \end{tabular}
    \includegraphics[width=0.156\textwidth]{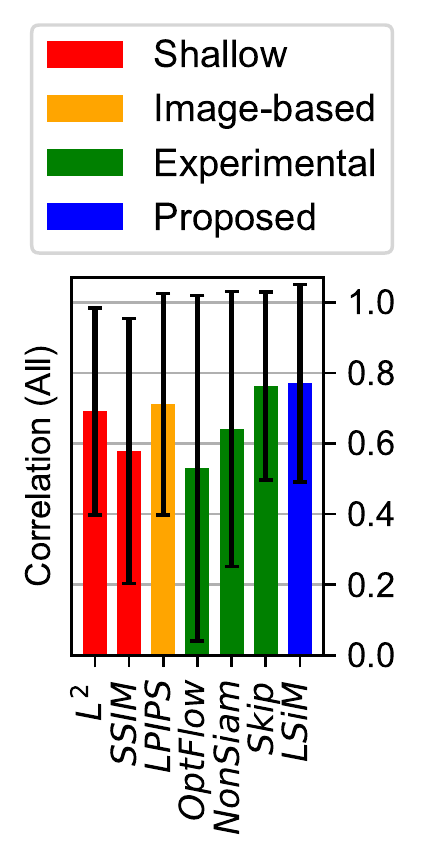}
\end{table*}

\begin{table*}[p]
    \vspace{-0.1cm}
    \caption{Performance comparison on test data sets measured by computing mean and std. dev. (in brackets) of Pearson correlation coefficients (ground truth against predicted distances) from individual data samples. \best{Bold+underlined} values show the best performing metric for each data set, \bestErr{bold} values are within a $0.01$ error margin of the best performing, and \bad{italic} values are $0.2$ or more below the best performing. }
    \label{table: results pearson mean 2}
    \vspace{0.1cm}
    
    \centering
    \small
    \begin{tabular}{l c | c c c c c}
        \toprule
        \multirow{2}{*}[-1.3mm]{\bf Metric} & \multicolumn{6}{c}{\bf Test data sets} \\
        \cmidrule(lr){2-7}
        & \texttt{TID} & \texttt{LiqN} & \texttt{AdvD} & \texttt{Sha} & \texttt{Vid} & \texttt{All} \\
        \cmidrule(lr){1-7}

        \it $L^2$                               & 0.84 (0.08) & 0.75 (0.18) & 0.57 (0.38) & \bad{0.67 (0.18)} & 0.84 (0.27) & 0.69 (0.29) \\
        \it SSIM                                & 0.81 (0.20) & \bad{0.26 (0.38)} & \best{0.71 (0.31)} & \bad{0.53 (0.32)} & 0.77 (0.28) & 0.58 (0.38) \\
        \it LPIPS v0.1.                         & \bestErr{0.87 (0.11)} & \bad{0.51 (0.34)} & 0.63 (0.34) & 0.85 (0.14) & \bestErr{0.87 (0.22)} & 0.71 (0.31) \\
        \cmidrule(lr){2-7}

        \it AlexNet\textsubscript{random}       & 0.84 (0.10) & 0.67 (0.24) & 0.65 (0.33) & 0.74 (0.18) & 0.85 (0.26) & 0.72 (0.28) \\
        \it AlexNet\textsubscript{frozen}       & 0.86 (0.11) & \bad{0.41 (0.37)} & 0.64 (0.34) & 0.87 (0.14) & \bestErr{0.87 (0.22)} & 0.70 (0.34) \\
        \it Optical flow                        & 0.74 (0.67) & \bad{0.50 (0.34)} & \bad{0.32 (0.53)} & \bad{0.63 (0.45)} & 0.78 (0.45) & \bad{0.53 (0.49)} \\
        \it Non-Siamese                         & 0.87 (0.12) & \best{0.84 (0.12)} & 0.66 (0.34) & \bad{0.31 (0.45)} & 0.83 (0.26) & 0.64 (0.39) \\
        \it Skip\textsubscript{from scratch}    & \bestErr{0.87 (0.12)} & 0.80 (0.16) & 0.63 (0.37) & 0.80 (0.17) & \best{0.87 (0.20)} & \bestErr{0.76 (0.27)} \\
        \cmidrule(lr){2-7}
        
        \it LSiM\textsubscript{noiseless}       & \bestErr{0.87 (0.11)} & \bad{0.64 (0.29)} & 0.60 (0.38) & 0.86 (0.15) & \bestErr{0.86 (0.22)} & 0.73 (0.31) \\
        \it LSiM\textsubscript{strong noise}    & 0.83 (0.12) & \bad{0.39 (0.38)} & 0.55 (0.36) & \best{0.91 (0.17)} & \bestErr{0.86 (0.25)} & 0.67 (0.37) \\
        \it LSiM (ours)                         & \best{0.88 (0.10)} & 0.81 (0.15) & 0.60 (0.37) & 0.88 (0.16) & 0.85 (0.23) & \best{0.77 (0.28)} \\
   
        \bottomrule
    \end{tabular}
\end{table*}

In the Tables~\ref{table: results pearson mean 1} and \ref{table: results pearson mean 2}, we employ a different, more intuitive approach to determine combined correlation values for each data set using the Pearson correlation. We are no longer analyzing the entire predicted distance distribution and the ground truth distribution at once as done above.
Instead, we individually compute the correlation between the ground truth and the predicted distances for the single data samples of the data set. From the single correlation values, we compute the mean and standard deviations shown in the tables. Note that this approach potentially produces less accurate comparison results, as small errors in the individual computations can accumulate to larger deviations in mean and standard deviation.
Still, both tables lead to very similar conclusions: The best performing metrics are almost the same, and low combined correlation values match with results that have a high standard deviation and a low mean.

\begin{figure*}[hb]
    \centering
    \includegraphics[width=1.0\textwidth]{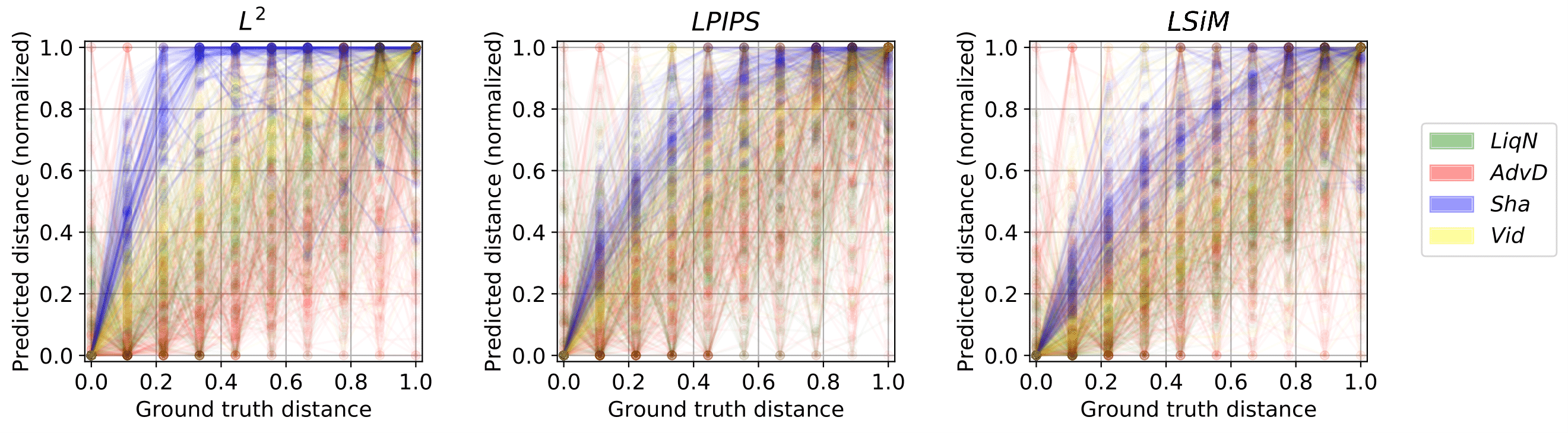} 
    \vspace{-0.7cm}
    \caption{Distribution evaluation of ground truth distances against normalized predicted distances for $L^2$, \textit{LPIPS} and \textit{LSiM} on all test data (color coded).}
    \label{fig: evaluation scatter}
\end{figure*}

\begin{figure*}[hb]
    \centering
    \includegraphics[width=1.0\textwidth]{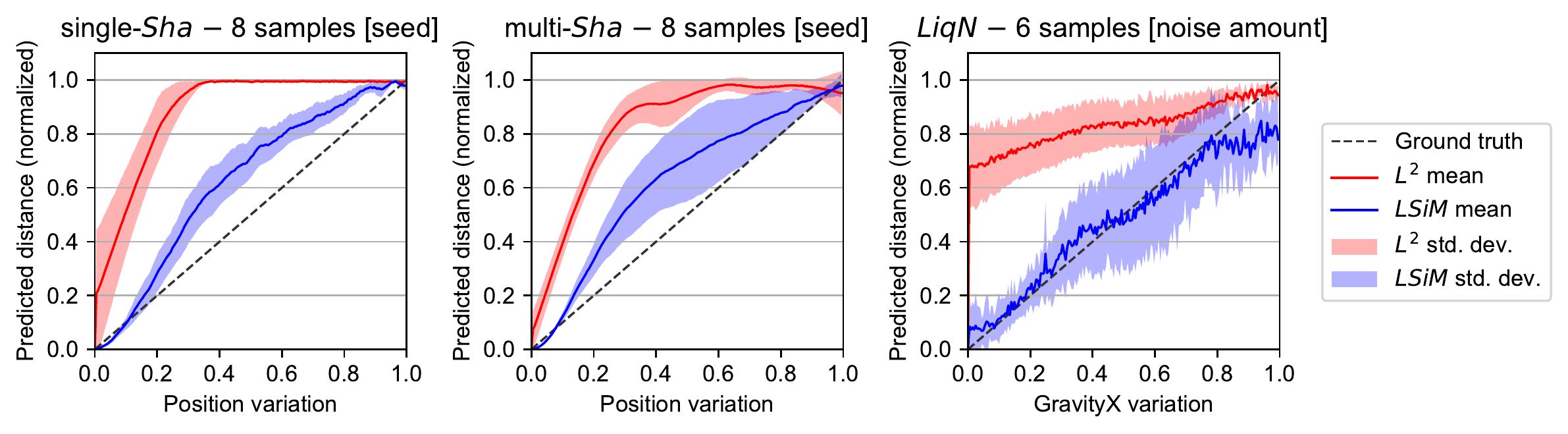}
    \vspace{-0.7cm}
    \caption{Mean and standard deviation of normalized distances over multiple data samples for $L^2$ and \textit{LSiM}. The samples differ by the quantity displayed in brackets. Each data sample uses 200 parameter variation steps instead of 10 like the others in our data sets. For the shape data the position of the shape varies and for the liquid data the gravity in x-direction is adjusted.}
    \label{fig: convergence study}
\end{figure*}

Fig.~\ref{fig: evaluation scatter} shows a visualization of predicted distances $\vc$ against ground truth distances $\vd$ for different metrics on every sample from the test sets. Each plot contains over 6700 individual data points to illustrate the global distance distributions created by the metrics, without focusing on single cases. A theoretical optimal metric would recover a perfectly narrow distribution along the line $\vc=\vd$, while worse metrics recover broader, more curved distributions.
Overall, the sample distribution of an $L^2$ metric is very wide. \textit{LPIPS} manages to follow the optimal diagonal a lot better, but our approach approximates it with the smallest deviations, as also shown in the tables above. The $L^2$ metric performs very poorly on the shape data indicated by the too steeply increasing blue lines that flatten after a ground truth distance of 0.3. \textit{LPIPS} already significantly reduces this problem, but \textit{LSiM} still works slightly better.

A similar issue is visible for the Advection-Diffusion data, where for $L^2$ a larger number of red samples is below the optimal $\vc=\vd$ line, than for the other metrics. \textit{LPIPS} has the worst overall performance for liquid test set, indicated by the large number of fairly chaotic green lines in the plot. On the video data, all three metrics perform similarly well.

A fine-grained distance evaluation in 200 steps of $L^2$ and our \textit{LSiM} metric via the mean and standard deviation of different data samples is shown in Fig.~\ref{fig: convergence study}. Similar to Fig.~\ref{fig: evaluation scatter}, the mean of an optimal metric would follow the ground truth line with a standard deviation of zero, while the mean of worse metrics deviates around the line with a high standard deviation. The plot on the left combines eight samples with different seeds from the \texttt{Sha} data set, where only a single shape is used. Similarly, the center plot aggregates eight samples from \texttt{Sha} with more than one shape. The right plot shows six data samples from the \texttt{LiqN} test set that vary by the amount of noise that was injected into the simulation.

The task of only tracking a single shape in the example on the left is the easiest of the three shown cases. Both metrics have no problem to recover the position change until a variation of 0.4, where $L^2$ can no longer distinguish between the different samples. Our metric recovers distances with a continuously rising mean and a very low standard deviation. The task in the middle is already harder, as multiple shapes can occlude each other during the position changes. Starting at a position variation of 0.4, both metrics have a quite high standard deviation, but the proposed method stays closer to the ground truth line. $L^2$ shows a similar issue as before because it flattens relatively fast. The plot on the right features the hardest task. Here, both metrics perform similar as each has a different problem in addition to an unstable mean. Our metric stays close to the ground truth, but has a quite high standard deviation starting at about a variation of 0.4. The standard deviation of $L^2$ is lower, but instead it starts off with a big jump from the first few data points. To some degree, this is caused by the normalization of the plots, but it still overestimates the relative distances for small variations in the simulation parameter.

These findings also match with the distance distribution evaluations in Fig.~\ref{fig: evaluation scatter} and the tables above: Our method has a significant advantage over shallow metrics on shape data, while the differences of both metrics become much smaller for the liquid test set.


\section{Notation} \label{append: notation}
In this work, we follow the notation suggested by \citeauthor{goodfellow2016}. Vector quantities are displayed in bold, and tensors use a sans-serif font. Double-barred letters indicate sets or vector spaces. The following symbols are used:

\bgroup
\def\arraystretch{1.5}
\begin{tabular}{p{2cm}p{5.4cm}}
$\displaystyle \sR$ & Real numbers \\
$\displaystyle i,j$ & Indexing in different contexts \\
$\displaystyle \sI$ & Input space of the metric, i.e., color images/field data of size $224\times224\times3$\\
$\displaystyle a$ & Dimension of the input space $\sI$ when flattened to a single vector\\
$\displaystyle \vx,\vy,\vz$ & Elements in the input space $\sI$ \\
$\displaystyle \sL$ & Latent space of the metric, i.e., sets of 3\textsuperscript{rd} order feature map tensors\\
$\displaystyle b$ & Dimension of the latent space $\sL$ when flattened to a single vector\\
$\displaystyle \tilde{\vx},\tilde{\vy},\tilde{\vz}$ & Elements in the latent space $\sL$, corresponding to $\vx,\vy,\vz$\\
$\displaystyle w$ & Weights for the learned average aggregation (1 per feature map) \\
$\displaystyle p_0, p_1, \dotsc$ & Initial conditions / parameters of a numerical simulation \\
$\displaystyle n$ & Number of variations of a simulation parameter, thus determines length of the network input sequence\\
\end{tabular}
\egroup

\bgroup
\def\arraystretch{1.5}
\begin{tabular}{p{2cm}p{5.4cm}}
$\displaystyle o_0, o_1, \dotsc, o_n$ & Series of outputs of a simulation with increasing ground truth distance to $o_0$\\
$\displaystyle \Delta$ & Amount of change in a single simulation parameter\\
$\displaystyle t_1, t_2, \dotsc, t_t$ & Time steps of a numerical simulation\\
$\displaystyle v$ & Variance of the noise added to a simulation\\
$\displaystyle \vc$ & Ground truth distance distribution, determined by the data generation via $\Delta$ \\
$\displaystyle \vd$ & Predicted distance distribution~ (supposed to match the corresponding $\vc$)\\
$\displaystyle \bar{\vc}, \bar{\vd}$ & Mean of the distributions $\vc$ and $\vd$\\
$\displaystyle \left\| \dotsc \right\|_2$ & Euclidean norm of a vector\\
$\displaystyle m(\vx,\vy)$ & Entire function computed by our metric\\
$\displaystyle m_1(\vx,\vy)$ & First part of $m(\vx,\vy)$, i.e., base network and feature map normalization\\
$\displaystyle m_2(\tilde{\vx},\tilde{\vy})$ & Second part of $m(\vx,\vy)$, i.e., latent space difference and the aggregations \\
$\displaystyle \tG$ & 3\textsuperscript{rd} order feature tensor from one layer of the base network \\
$\displaystyle g_b,g_c,g_x,g_y$ & Batch ($g_b$), channel ($g_c$), and spatial dimensions ($g_x,g_y$) of $\tG$\\
$\displaystyle f$ & Optical flow network\\
$\displaystyle f^{\vx\vy}, f^{\vy\vx}$ & Flow fields computed by an optical flow network $f$ from two inputs in $\sI$\\
$\displaystyle f^{\vx\vy}_1, f^{\vx\vy}_2$ & Components of the flow field $f^{\vx\vy}$\\
$\displaystyle \nabla, \nabla^2$ & Gradient ($\nabla$) and Laplace operator ($\nabla^2$)\\
$\displaystyle \partial$ & Partial derivative operator\\
$\displaystyle t$ & Time in our PDEs\\
$\displaystyle u$ & Velocity in our PDEs\\
$\displaystyle \nu$ & Kinematic viscosity / diffusion coefficient in our PDEs\\
$\displaystyle d, \rho$ & Density in our PDEs\\ 
\end{tabular}
\egroup

\bgroup
\def\arraystretch{1.5}
\begin{tabular}{p{2cm}p{5.4cm}}
$\displaystyle P$ & Pressure in the Navier-Stokes Equations\\
$\displaystyle g$ & Gravity in the Navier-Stokes Equations\\
\end{tabular}
\egroup

\end{document}